\documentclass[preprint,review]{elsarticle}

\usepackage{lineno,hyperref}
\modulolinenumbers[5]

\usepackage{multirow}
\usepackage{amssymb}
\usepackage{amsmath}
\usepackage{graphicx}
\usepackage{subfigure}
\usepackage{epsfig}
\usepackage{makeidx}
\usepackage{mathtools}
\usepackage{multicol}
\usepackage{theorem}
\usepackage{enumerate}
\usepackage{algorithm}
\usepackage{algorithmic}
\usepackage{subeqnarray}
\usepackage{mathrsfs}
\usepackage{natbib}
\usepackage{booktabs}
\usepackage{epstopdf}
\usepackage{bm}
\usepackage{natbib}
\usepackage{rotating}
\usepackage{hyperref}
\hypersetup{
    colorlinks=true,
    linkcolor=blue,
    filecolor=blue,
    urlcolor=blue,
    citecolor=cyan,
}

\begin{document}

\begin{frontmatter}

\title{Transformed $\ell_1$ Regularization for Learning Sparse Deep Neural Networks}

\author[Mathe]{Rongrong Ma}
\ead{marongrong16@mails.ucas.ac.cn}

\author[Infor]{Jianyu Miao}
\ead{jymiao@haut.edu.cn}

\author[Eco]{Lingfeng Niu\corref{cor1}}
\ead{niulf@ucas.ac.cn}

\author[Ant]{Peng Zhang}
\ead{zhangpeng04@gmail.com}

\cortext[cor1]{Corresponding author}

\address[Mathe]{School of Mathematical Sciences, University of Chinese Academy of Sciences, Beijing, 100049, China}
\address[Infor]{College of Information Science and Engineering, Henan University of Technology, Zhengzhou, 450001, China}
\address[Eco]{School of Economics and Management, University of Chinese Academy of Sciences, Beijing, 100190, China}
\address[Ant]{Ant Financial Services Group, Hangzhou, 310012, China}






\begin{abstract}
Deep neural networks (DNNs) have achieved extraordinary success in numerous areas. However, to attain this success, DNNs often carry a large number of weight parameters, leading to heavy costs of memory and computation resources. Overfitting is also likely to happen in such network when the training data are insufficient. These shortcomings severely hinder the application of DNNs in resource-constrained platforms. In fact, many network weights are known to be redundant and can be removed from the network without much loss of performance. To this end, we introduce a new non-convex integrated transformed $\ell_1$ regularizer to promote sparsity for DNNs, which removes both redundant connections and unnecessary neurons simultaneously. To be specific, we apply the transformed $\ell_1$ to the matrix space of network weights and utilize it to remove redundant connections. Besides, group sparsity is also employed as an auxiliary to remove unnecessary neurons. An efficient stochastic proximal gradient algorithm is presented to solve the new model at the same time. To the best of our knowledge, this is the first work to utilize a non-convex regularizer in sparse optimization based method to promote sparsity for DNNs. Experiments on several public datasets demonstrate the effectiveness of the proposed method.
\end{abstract}


\end{frontmatter}

\section{Introduction}
Recently, deep neural networks (DNNs) have achieved remarkable success in many fields~\cite{Lecun2015Deep,schmidhuber2015deep,goodfellow2016deep,deng2014deep}.
One of the key factors in this success is its expressive power, which heavily relies on the large number of parameters~\cite{zhou2016less,alvarez2016learning,yoon2017combined}. For example, VGG~\cite{simonyan2014very}, which is a convolutional neural network and wins the ImageNet Large Scale Visual Recognition Challenge 2014, consists of 15M neurons and up to 144M parameters. Increased number of parameters means increasing burdens on both memory and computation power, which make DNNs costly for training and inapplicable to resource limited platforms~\cite{alvarez2016learning}. Moreover, models with massive parameters are more easily overfitting when the training data are insufficient~\cite{yoon2017combined,scardapane2017group}. These challenges seriously hinder the further application of DNNs~\cite{alvarez2016learning}. However, DNNs are known to have many redundant parameters \cite{alvarez2016learning,yoon2017combined,scardapane2017group,denil2013predicting,Cheng2015An}. For example, \cite{denil2013predicting} shows that in some networks, only $5\%$ of parameters are enough to achieve acceptable models. A number of research works have focused on compressing and accelerating DNNs~\cite{alvarez2016learning,cheng2017survey,han2015learning,hinton2012improving,cheng2018model}. Among these techniques, one class pays attention to promote sparsity in DNNs.

We classify the existing works on sparsity promotion for DNNs into three categories: pruning, dropout, and the sparse optimization based method. Pruning removes weight parameters which are insensitive to the performance with respect to established dense networks. The seminal work is the Biased Weight Decay~\cite{hanson1989comparing}. Then, the works~\cite{hassibi1993second,Cun1989Optimal,hassibi1993optimal} use the Hessian loss function to remove network connections. In a recent work~\cite{han2015learning}, connections having slight effect are removed to obtain sparse networks. There are also methods using various criteria to determine which parameters or connections are unnecessary~\cite{narang2017exploring,anwar2017structured}. However, in these approaches, the pruning criteria require manual setups of layer sensitivity and heuristic assumptions are also necessary during the pruning phase~\cite{cheng2017survey}.

Dropout reduces the size of networks during training by randomly dropping units along with their connections from DNNs~\cite{hinton2012improving,wan2013regularization,srivastava2014dropout}. Biased Dropout and Crossmap Dropout~\cite{poernomo2018biased} are proposed to implement dropout on hidden units and convolutional layers respectively. These methods can reduce overfitting efficiently and improve the performance. Nonetheless, training a Dropout network usually takes more time than training a standard neural network, even they are with the same architecture~\cite{srivastava2014dropout}. In addition, dropout can only simplify networks during training. Full-sized networks are still needed in prediction phase.

The sparse optimization based method promotes sparsity in networks by introducing structured sparse regularization term into the optimization model of DNN, and zeroing out the redundant parameters during the process of training. Compared with pruning, this type of approaches do not rely on manual setups. In contrast to dropout, the simplified networks obtained by sparse optimization can also be used in the prediction stage. Moreover, different from most existing methods which compress network with negligible drop of accuracy, experiments show that some sparse optimization based methods could even achieve better performance than their original networks~\cite{alvarez2016learning,scardapane2017group,yoon2017combined}. Considering all these merits, we would construct sparse neural networks in the framework of sparse optimization.

The sparse optimization method can be utilized to various tasks to produce sparse solutions. The key challenge of this approach is the design of regularization functions. $\ell_0$ norm, which counts the number of non-zero elements, is the most intuitive form of sparse regularizers and can promote the sparsest solution. However, minimizing $\ell_0$ problem is combinatory and usually NP-hard~\cite{natarajan1995sparse}. The $\ell_1$ norm is the most commonly used surrogate, which is convex and can be solved easily. Although $\ell_1$ enjoys several good properties, it is sensitive to outliers and may cause serious bias in estimation~\cite{Fan1999Variable,fan2001variable}. To overcome this defect, many non-convex surrogates are proposed and analyzed, including smoothly clipped absolute deviation (SCAD)~\cite{Fan1999Variable}, log penalty~\cite{candes2008enhancing,mazumder2011sparsenet}, capped $\ell_1$~\cite{zhang2009multi,zhang2010analysis}, minimax concave penalty (MCP)~\cite{zhang2010nearly}, $\ell_p$ penalty with $p \in(0,1)$~\cite{Xu2010Data,Xu2012L1,Krishnan2009Fast}, the difference of $\ell_1$ and $\ell_2$ norms~\cite{esser2013method,lou2015computing,yin2015minimization} and transformed $\ell_1$~\cite{nikolova2000local,zhang2017minimization,zhang2018minimization}. More and more works have shown the good performance of non-convex regularizers in both theoretical analyses and practical applications. Generally speaking, non-convex regularizers are more likely to produce unbiased model with sparser solution. However, to the best of our knowledge, no work has applied sparse optimization methods with non-convex regularization to DNNs to promote sparsity. Therefore, we would like to introduce non-convex regularizer to the training model of neural networks in this work.

When applied in DNNs, sparse regularizer is supposed to zero out redundant weights and thus remove unnecessary connections. Since the variables in DNNs are weights, which are usually modeled as matrices or even tensors, we would like to employ a proper regularizer that can avoid augmenting excessive computation complexity. After considering the properties of common used non-convex regularizers, we choose transformed $\ell_1$ as the regularizer in our model. It satisfies the three desired properties that a regularizer should result in an estimator with, i.e. unbiasedness, sparsity and continuity~\cite{fan2001variable}. In addition, its thresholding function has a closed-form solution. In order to further minify the scale of the network, we also consider group sparsity as an auxiliary of transformed $\ell_1$ to remove unnecessary neurons
because of its remarkable performance in promoting neuron-level sparsity~\cite{zhou2016less,scardapane2017group,lebedev2016fast,fang2015graph,yuan2006model,simon2013sparse}. By combining the transformed $\ell_1$ and group sparsity together, we propose a new integrated transformed $\ell_1$  regularizer. Extensive experiments are carried out to show the effectiveness of our method. The contribution of this paper is three-folded:

\begin{itemize}
\item To obtain sparse DNNs, a new model with non-convex regularizer is proposed. The regularizer integrates transformed $\ell_1$ and group sparsity together. To the best of our knowledge, this is the first work which utilizes a non-convex regularizer in sparse optimization based method for DNNs.

\item To train the new model, an algorithm based on proximal gradient descent is proposed. Although the transformed $\ell_1$ is non-convex, the proximal operators in our algorithm have closed-form solutions and can be computed easily.

\item Experiments in computer vision are executed on several public datasets. Compared with three prominent baselines, experimental results show the effectiveness of the proposed regularizer.
\end{itemize}

The rest of the paper is organized as follows. Section 2 surveys existing sparse optimization based works which aim to promote sparsity in DNNs and some popular non-convex regularizers. Section 3 introduces the new integrated transformed $\ell_1$ regularizer and proposes a proximal gradient algorithm to deal with the new model at the same time. Experiments on several public classification datasets are reported in Section 4. We conclude the paper in Section 5.

\section{Related Work}
\subsection{Sparse optimization for DNNs}
Sparse optimization based approaches in DNNs achieve sparsity through introducing sparse regularization term to the objective function and turning the training process into an optimization problem. Some pruning methods are also equipped with an objective function regularized by some norms. However, these two categories of methods are inherently different. Pruning methods do not aim to learn the final values of the weights, but rather learn which connections are significant. In contrast, the final value of weights is the key criterion to remove connections in sparse optimization based approaches. Only the weights which are exactly zero will be regarded as uninformative ones and be further removed from the network.

In~\cite{collins2014memory}, sparse regularizers including the $\ell_1$ regularizer, the shrinkage operator and the projection to $\ell_0$ balls are applied to both convolutional layers and fully-connected layers in convolutional neural networks. Nevertheless, these methods often achieve sparsity at the expense of accuracy. \cite{zhou2016less} employs two sparse constraints, including tensor low rank constraint and group sparsity, to zero out weights. Group sparsity and $\ell_1$ norm are combined together in \cite{alvarez2016learning} to zero out redundant connections and achieve sparsity of the network. The work \cite{scardapane2017group}, which exploits the similar regularization as \cite{alvarez2016learning}, divides outgoing connections of each input neuron, outgoing connections of each neuron in hidden layer and biases into different groups and promote group-level sparsity. Group sparsity and exclusive sparsity are combined as a regularization term in a recent work~\cite{yoon2017combined} to enforce sparsity, by utilizing the sharing and competing relationships among various network weights. These methods can achieve sparsity with comparable or even better accuracy than the original network.

\subsection{Non-convex regularization function}
\begin{figure*}
	\centering
	\subfigure[$\ell_0$]{
		\label{Fig.sub.0}
		\includegraphics[width=0.3\textwidth]{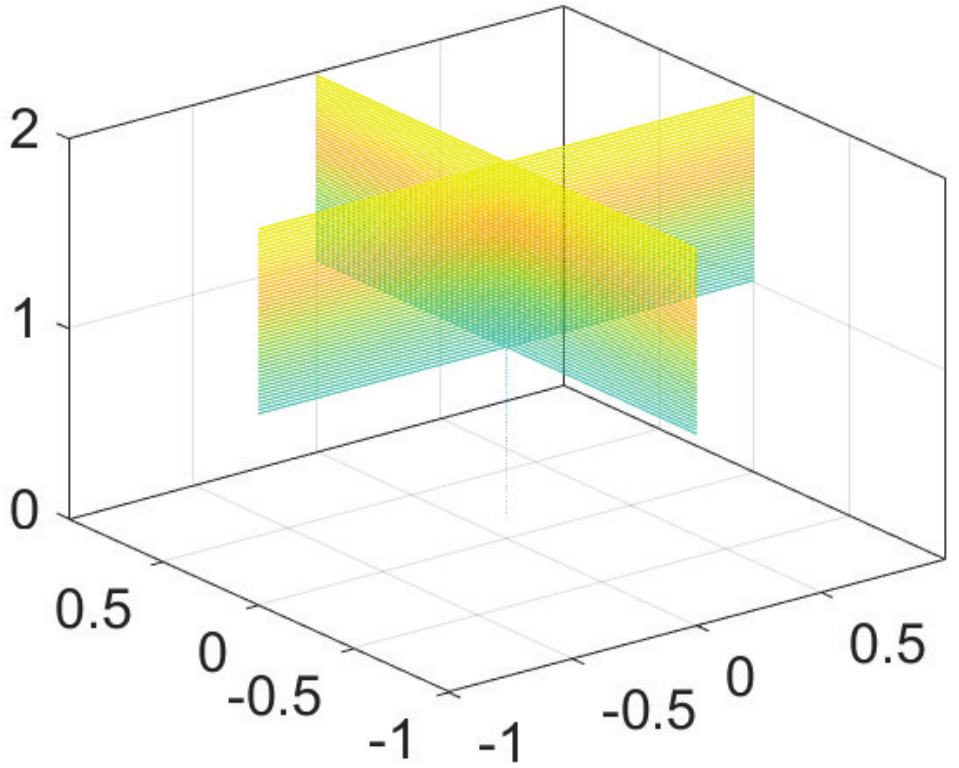}}
	\subfigure[$\ell_1$]{
		\label{Fig.sub.5}
		\includegraphics[width=0.3\textwidth]{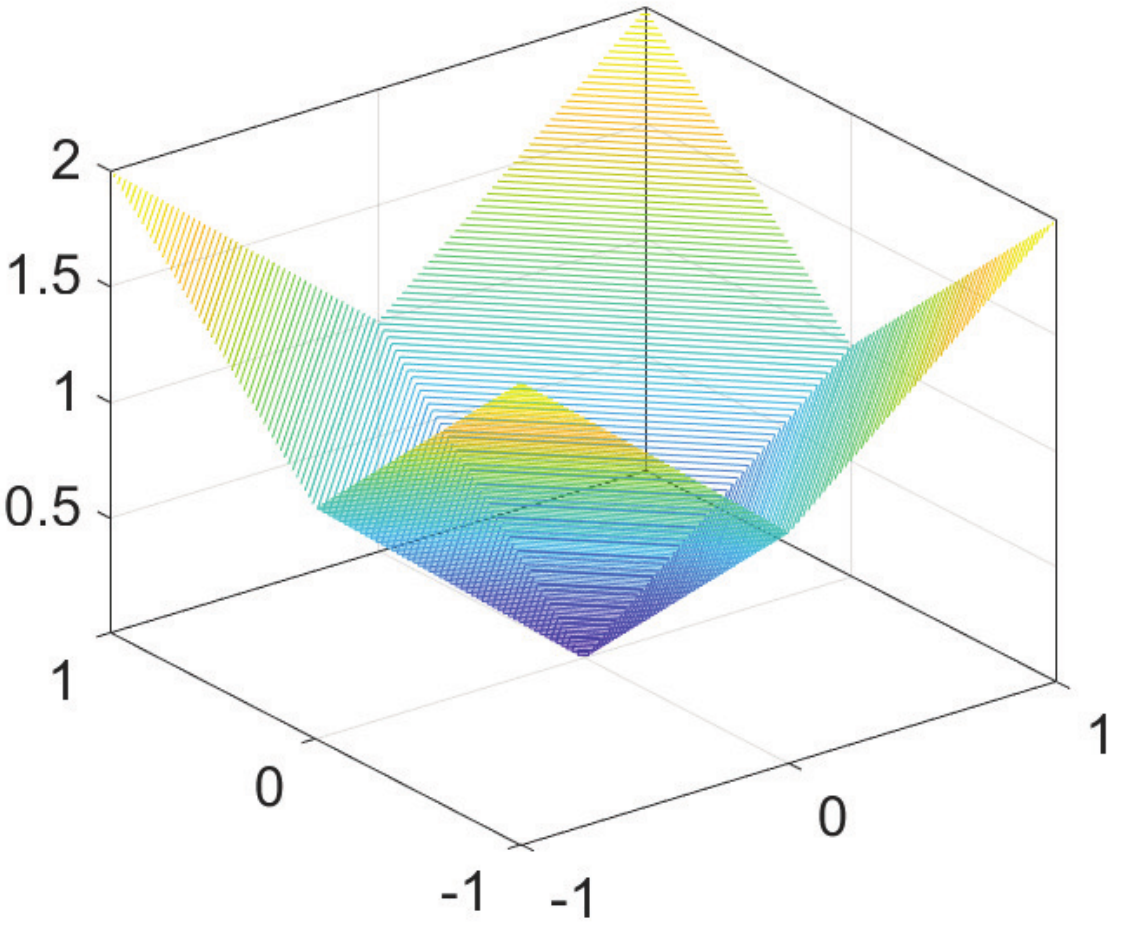}}
	\subfigure[SCAD]{
		\label{Fig.sub.11}
		\includegraphics[width=0.3\textwidth]{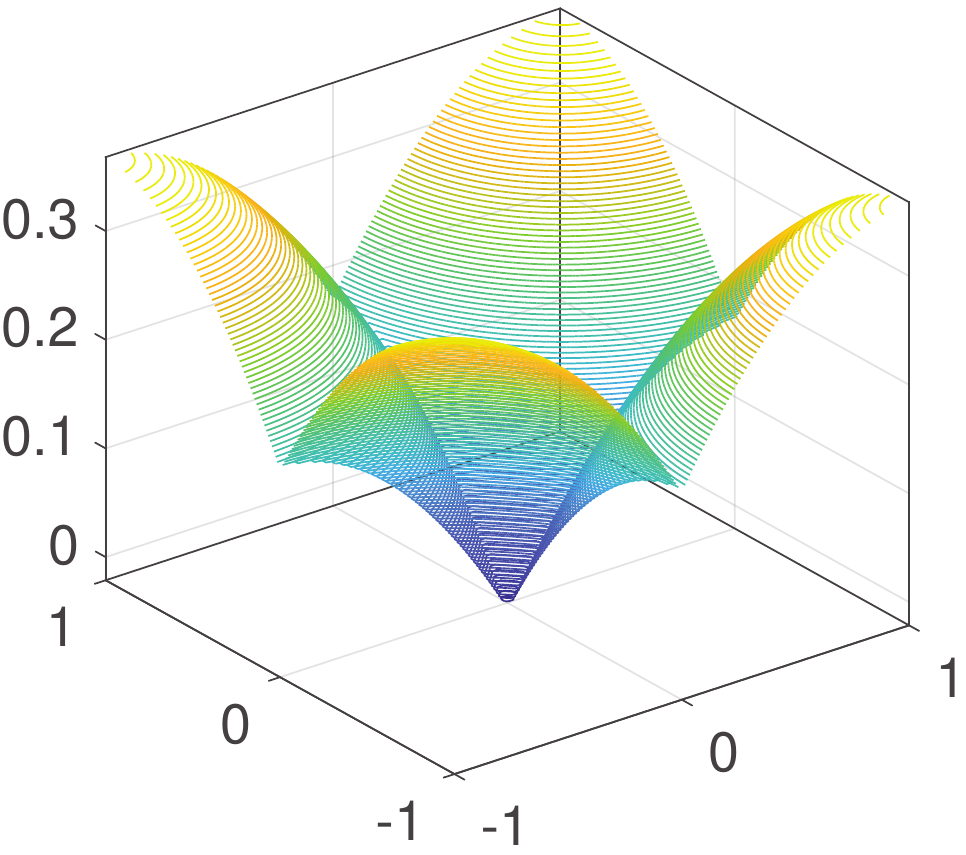}}
	\subfigure[MCP]{
		\label{Fig.sub.12}
		\includegraphics[width=0.3\textwidth]{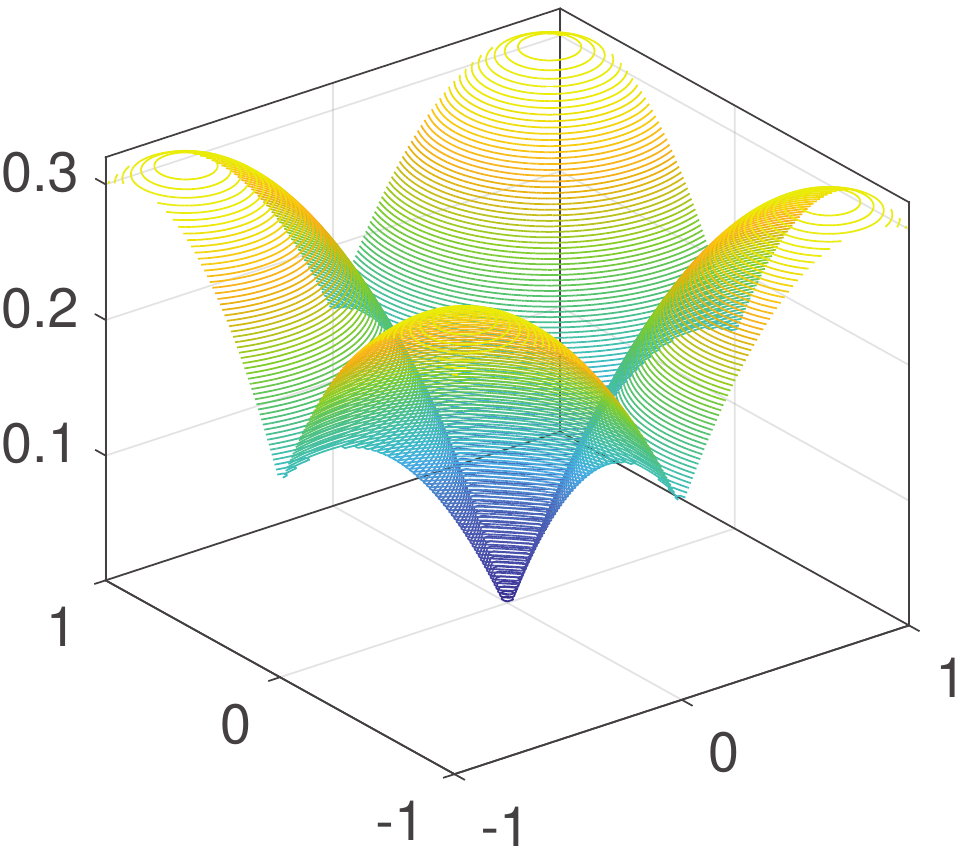}}
	\subfigure[capped $\ell_1$]{
		\label{Fig.sub.16}
		\includegraphics[width=0.3\textwidth]{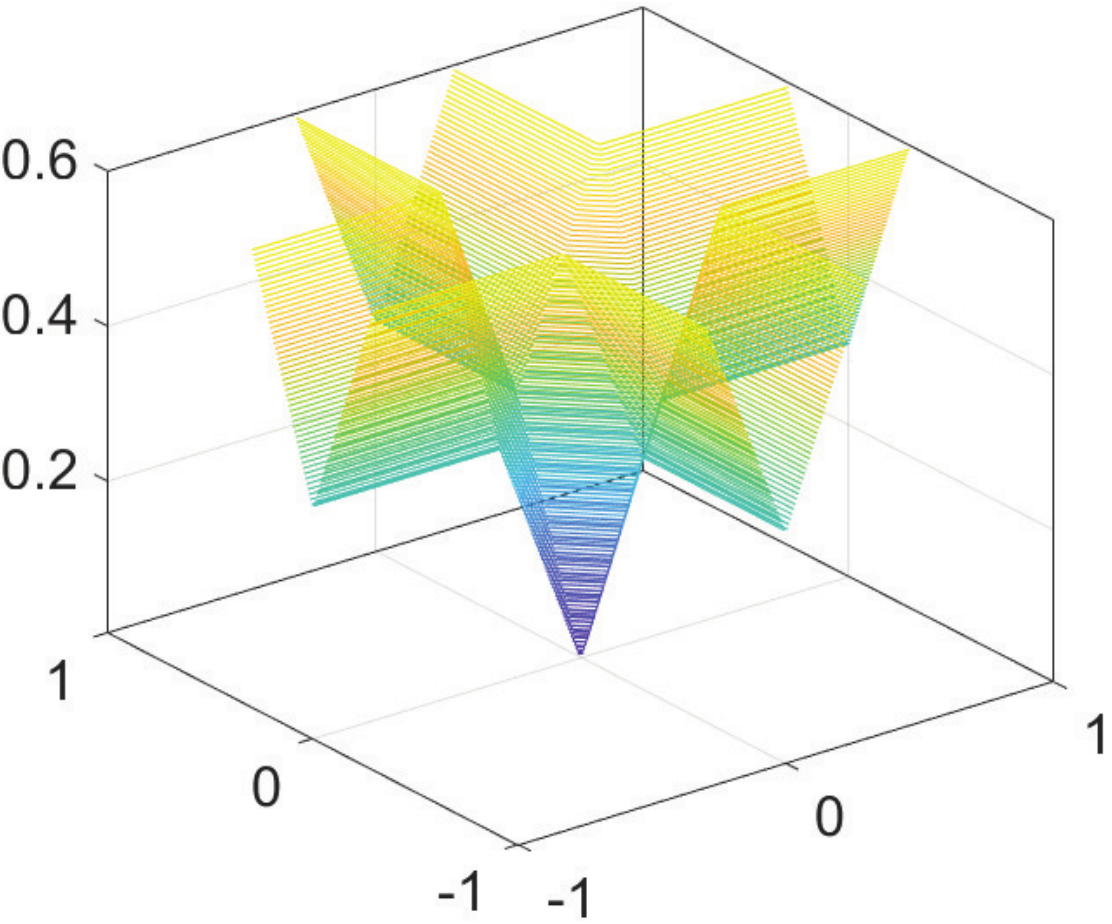}}
	\subfigure[log penalty]{
		\label{Fig.sub.15}
		\includegraphics[width=0.3\textwidth]{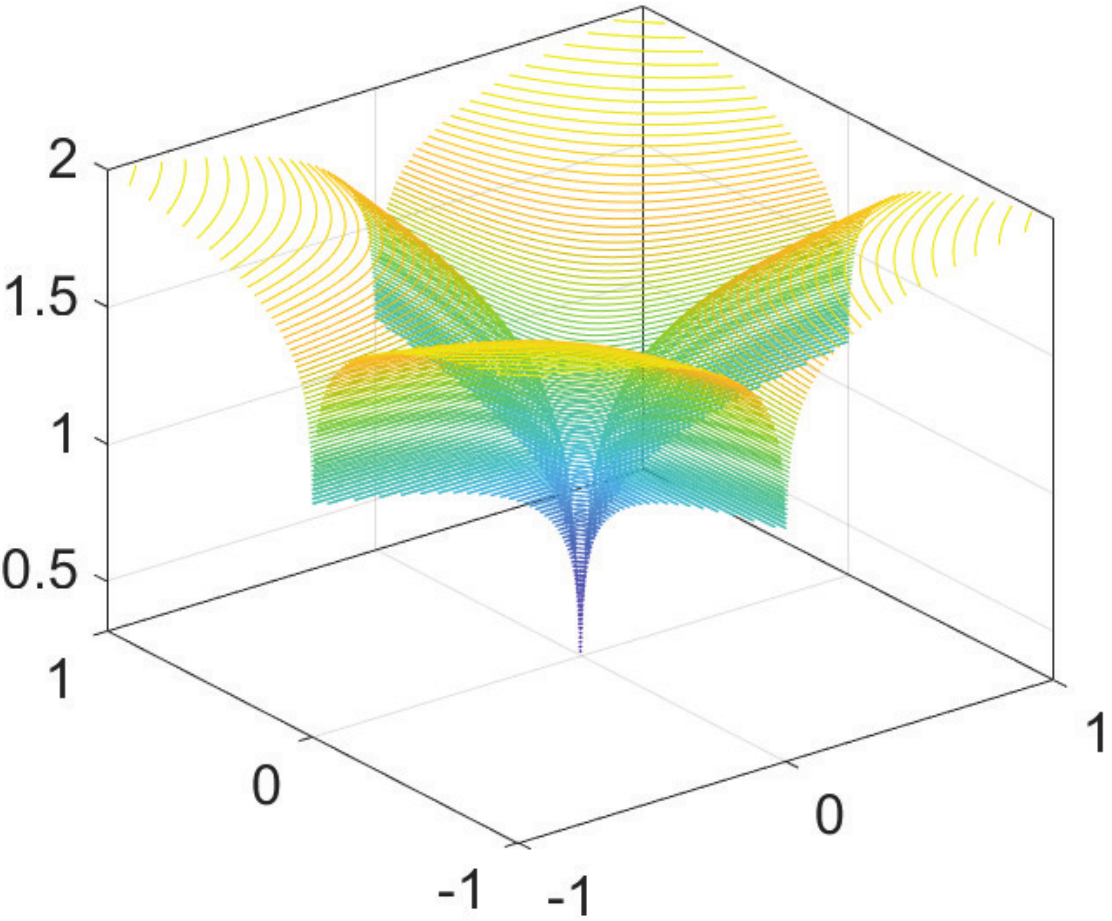}}
	\subfigure[$\ell_p$ with $p=1/2$]{
		\label{Fig.sub.10}
		\includegraphics[width=0.3\textwidth]{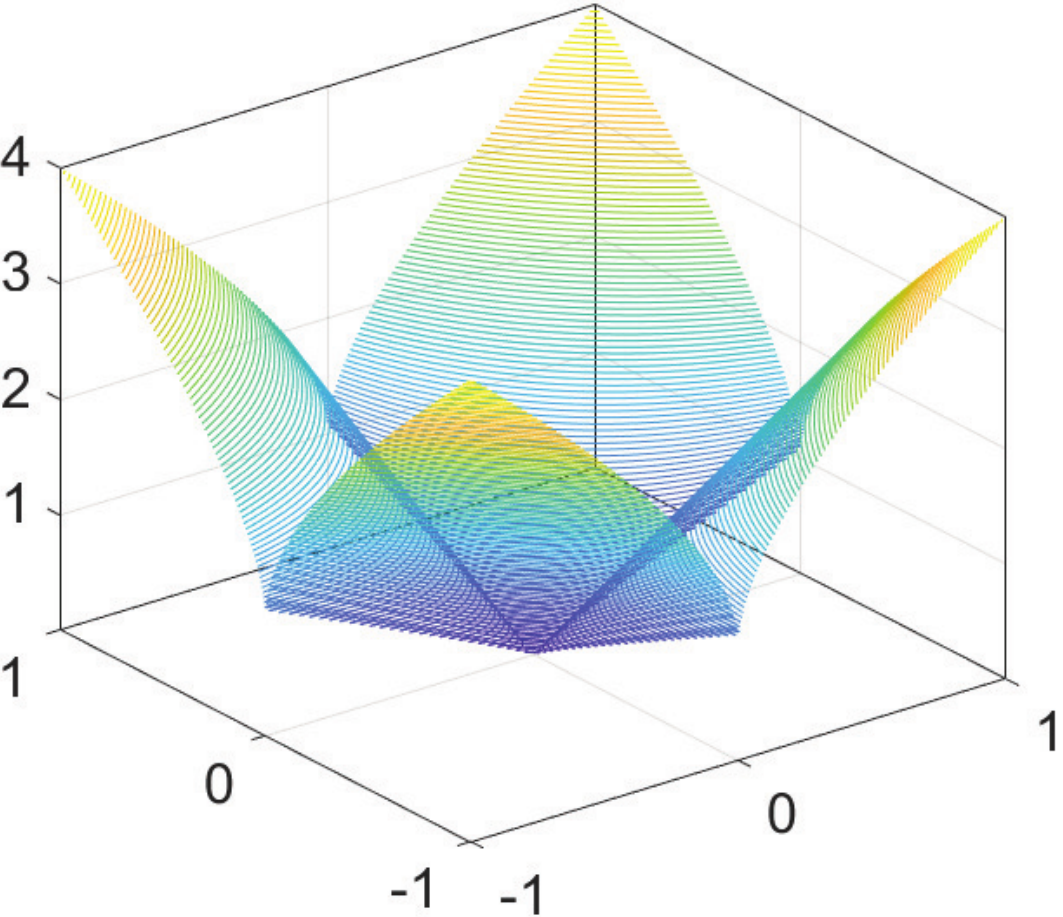}}
	\subfigure[$\ell_{1-2}$]{
		\label{Fig.sub.13}
		\includegraphics[width=0.3\textwidth]{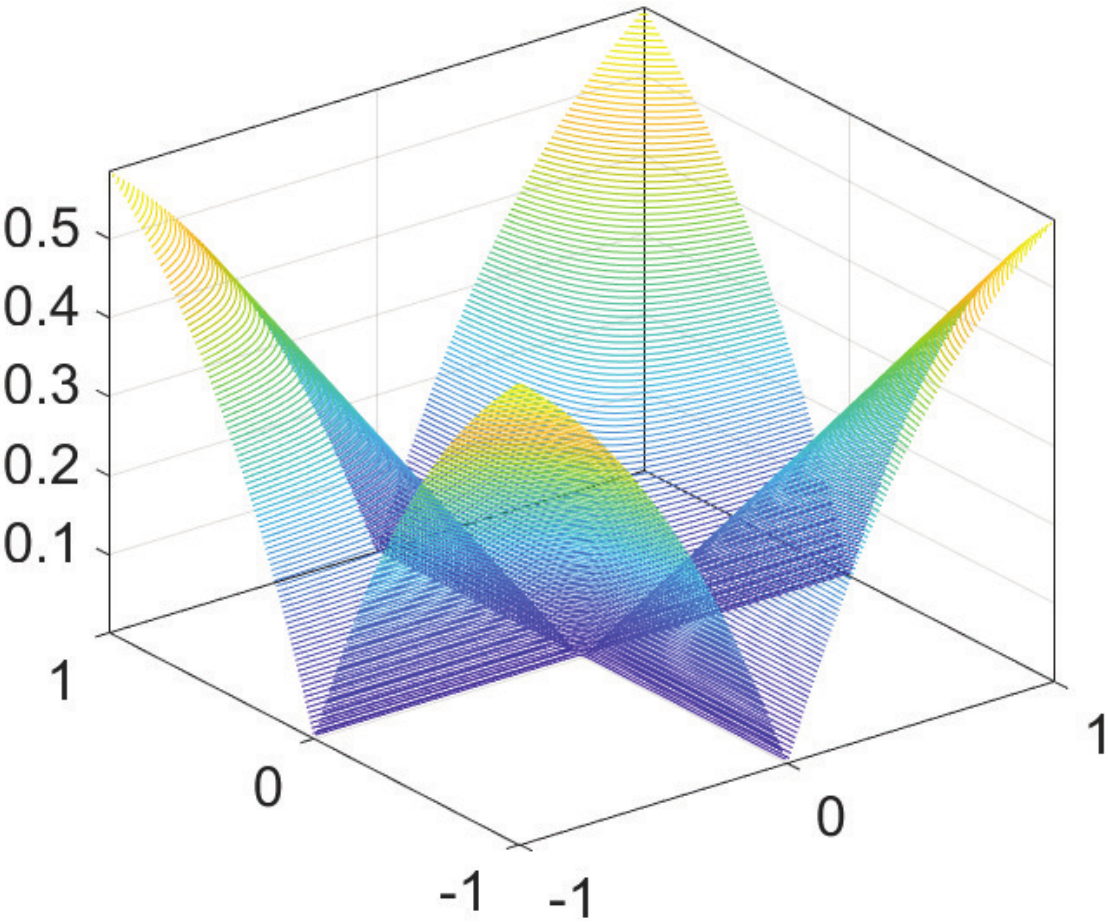}}
	\subfigure[transformed $\ell_1$ with $a=1$]{
		\label{Fig.sub.14}
		\includegraphics[width=0.3\textwidth]{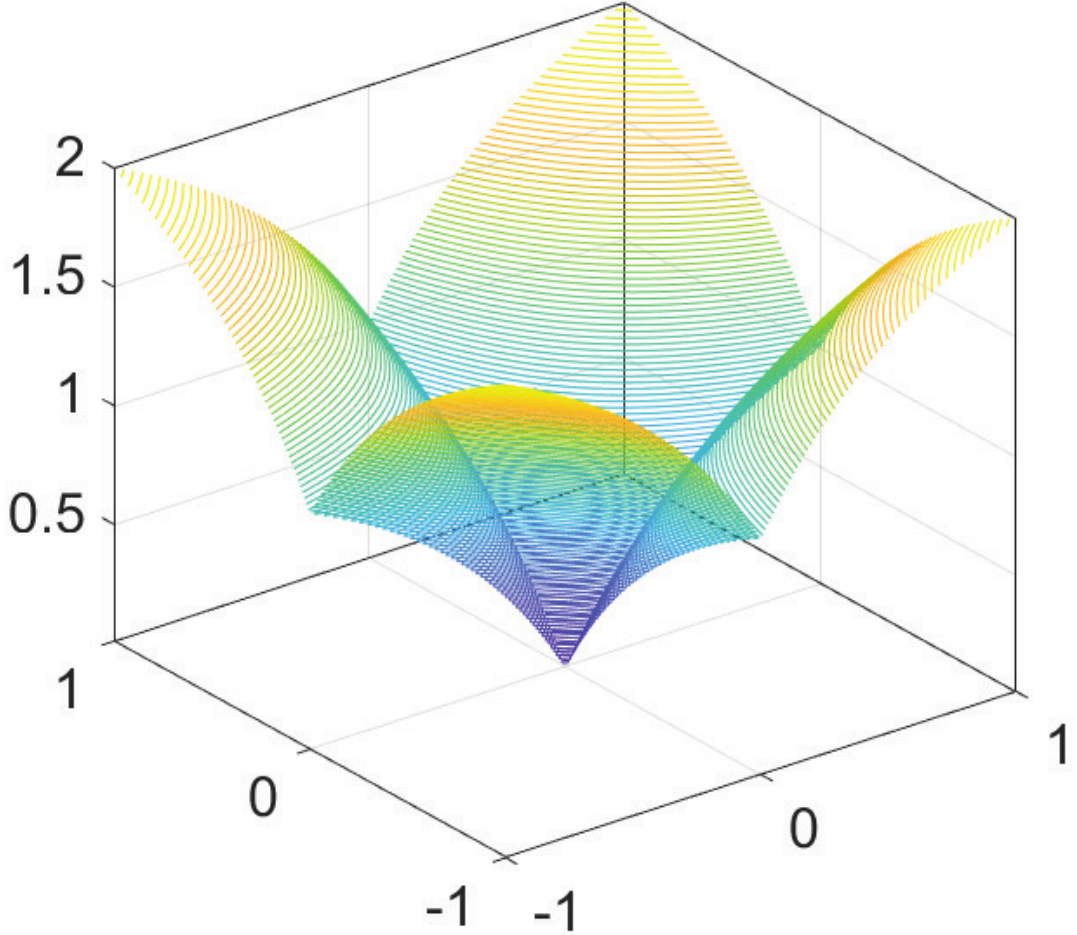}}
	\caption{\textbf{Contour plots for several popular convex and non-convex norms in two dimensions.} (a) $\ell_0$. (b) $\ell_1$. (c) SCAD with parameters $\lambda = 0.28$ and $a=3.7$. (d) MCP with $\lambda=0.4$ and parameter $\gamma=2$.  (e) capped $\ell_1$ with $a = 0.3$. (f) log penalty with $\gamma=10^3$. (g) $\ell_p$ with $p=1/2$. (h) $\ell_{1-2}$. (i) transformed $\ell_1$ with $a=1$. }	
	\label{nonconvexnorm}
\end{figure*}

The work \cite{fan2001variable} has proposed that a good penalty function which serves as the regularizer should result in an estimator with three desired properties: unbiasedness, sparsity and continuity. Obviously, the regularizers with these three properties simultaneously should be non-convex. The smoothly clipped absolute deviation (SCAD)~\cite{Fan1999Variable} is the first regularizer proven to fulfill these properties~\cite{fan2001variable}, whose definition for vector variable $\mathbf{x}=\{x_1,x_2,...,x_n\}\in \mathbb{R}^n$ is given as $\mathbf{P}(\mathbf{x};\lambda,\gamma)=\sum_{i=1}^{n}P(x_i;\lambda,\gamma)$, in which
\begin{eqnarray}\label{SCAD}
P(x_i;\lambda,\gamma)=
\left\{
\begin{array}{llll}
\lambda\vert x_i\vert,                                           &if \ \vert x_i\vert\leq \lambda \\
\frac{2\gamma\lambda\vert x_i\vert-x_i^2-\lambda^2}{2(\gamma-1)},  &if \  \lambda < \vert x_i\vert<\gamma\lambda\\
{\lambda^2(\gamma+1)}/2,                                 &if\ \vert x_i\vert\geq \gamma\lambda,
\end{array}
\right.
\end{eqnarray}
where $\lambda>0$ and $\gamma>2$. It is obvious that SCAD is a two-parameter function composed of three parts. Later, a single-parameter concave regularizer with two pieces, called minimax concave penalty (MCP), is proposed in \cite{zhang2010nearly}. It is formulated as $\mathbf{P}_\gamma(\mathbf{x};\lambda)=\sum_{i=1}^{n}P_\gamma(x_i;\lambda)$, where
\begin{eqnarray}\label{MCP}
P_\gamma(x_i;\lambda)=
\left\{
\begin{array}{lllll}
\lambda\vert x_i\vert -{x_i^2}/(2\gamma), &if \ \vert x_i\vert\leq\gamma\lambda\\
\gamma\lambda^2/2,               &if \ \vert x_i\vert>\gamma\lambda
\end{array}
\right.
\end{eqnarray}
for parameter $\gamma>1$. Log penalty is a generalization of elastic net family, which is formulated as $\mathbf{P}(\mathbf{x};\gamma)=\sum_{i=1}^{n}P(x_i,\gamma)$ with
\begin{align}
P(x_i;\gamma)=\frac{\log(\gamma \vert x_i\vert+1)}{\log(\gamma+1)},
\end{align}
where parameter $\gamma>0$. Through this penalty family, the entire continuum of penalties from $\ell_1$ ($\gamma \rightarrow 0_{+}$) to $\ell_0$ ($\gamma \rightarrow \infty$) can be obtained \cite{mazumder2011sparsenet}. Capped $\ell_1$ is another approximation of $\ell_0$ \cite{zhang2009multi}, whose definition is
\begin{align}
\mathbf{P}(\mathbf{x};a)=\sum_{i=1}^{n}\min(\vert x_i\vert,a),
\end{align}
where $a$ is a positive capped parameter. Obviously, when $a\rightarrow0$, $ \sum_i\min(\vert x_i\vert,a)/a\rightarrow\|\mathbf{x}\|_0$. Transformed $\ell_1$, which is a smooth version of capped $\ell_1$, is discussed in the works \cite{nikolova2000local,zhang2017minimization,zhang2018minimization}. Some other non-convex metrics with concise form are also considered as alternatives to improve $\ell_1$, including $\ell_p$ with $p \in(0,1)$~\cite{Xu2010Data,Xu2012L1,Krishnan2009Fast}, whose formula is
\begin{align}
\| \mathbf{x}\|_p = \left(\sum_{i=1}^{n} \vert x_i\vert^p\right)^{1/p},
\end{align}
and $\ell_{1-2}$~\cite{esser2013method,lou2015computing,yin2015minimization}, which is the difference between $\ell_1$ and $\ell_2$ norm. Contour plots of several popular regularizers are displayed in Fig.~\ref{nonconvexnorm}.

\section{DNNs with Transformed $\ell_1$ Regularizer}
Our objective is to construct a sparse neural network with less number of parameters and comparable or even better performance than the dense model. In neural networks with multiple layers, let $W^{(l)}$ represent the weight matrix of $l$-th layer. By regularizing the weights of each layer respectively, the training objective function for supervised learning can be formulated as
\begin{align}\label{objec}
\min_{\{W^{(l)}\}}\mathcal{L}\left(\{W^{(l)}\},\mathcal{T}\right)+\lambda\sum_{l=1}^{L}\Omega(W^{(l)}),
\end{align}
where $\mathcal{T}=\left\{x_i,y_i\right\}_{i=1}^N$ is a training dataset that has $N$ instances, in which $x_i\in \mathbb{R}^p$ is a $p$-dimension input sample and $y_i\in \left\{1,...,K\right\}$ is its corresponding class label. $\lambda$ is a positive hyperparameter, which controls the balance between the loss term $\mathcal{L}(\{W^{(l)}\},\cdot)$ and the regularization term $\sum_{l=1}^{L}\Omega(W^{(l)})$. To induce sparsity in DNNs effectively, we will concentrate on constructing a proper sparse regularization function $\Omega(W^{(l)})$ in this paper.

As pointed out in \cite{yoon2017combined}, sparse regularizer can promote different weights at each layer to compete for few significant features, resulting in those weights remaining fitting to different features as much as possible and thus reducing the dependence and redundance among them. Therefore, this sparsity-inducing problem can also be considered from the respect of feature selection. Although the regularization function in the feature selection problem can be both convex and non-convex~\cite{gui2017feature}, it has been shown that non-convex regularizers outperform convex ones in numerous tasks~\cite{bradley1998feature,xiang2013efficient,zhang2014feature,shi2018feature}. Therefore, we would like to seek an appropriate non-convex regularizer to promote sparsity in DNNs.

\subsection{The model}
The transformed $\ell_1$ (T$\ell_1$) functions are a one parameter family of bilinear transformations composed with the absolute value function~\cite{nikolova2000local,zhang2017minimization,zhang2018minimization}. Mathematically, the T$\ell_1$ function for a scalar variable $x$ is defined as follows,
\begin{align}\label{TL122}
\rho_a(x)=\frac{(a+1)\left|x\right|}{a+\left|x\right|},
\end{align}
where $a$ is a positive parameter which controls the shape of the function. One can easily verify that when $a$ approaches zero, $\rho_a(x)$ tends to an indicative function $I(x)$ whose definition is: $I(x)=1$, if $x\neq0$ and $I(x)=0$, otherwise. In contrast, when $a$ approaches infinity, $\rho_a(x)$ tends to the absolute value function $\left|x\right|$ .

\begin{figure}
	\setlength{\abovecaptionskip}{-0.5cm}
	\centering
	\includegraphics[width=1\textwidth]{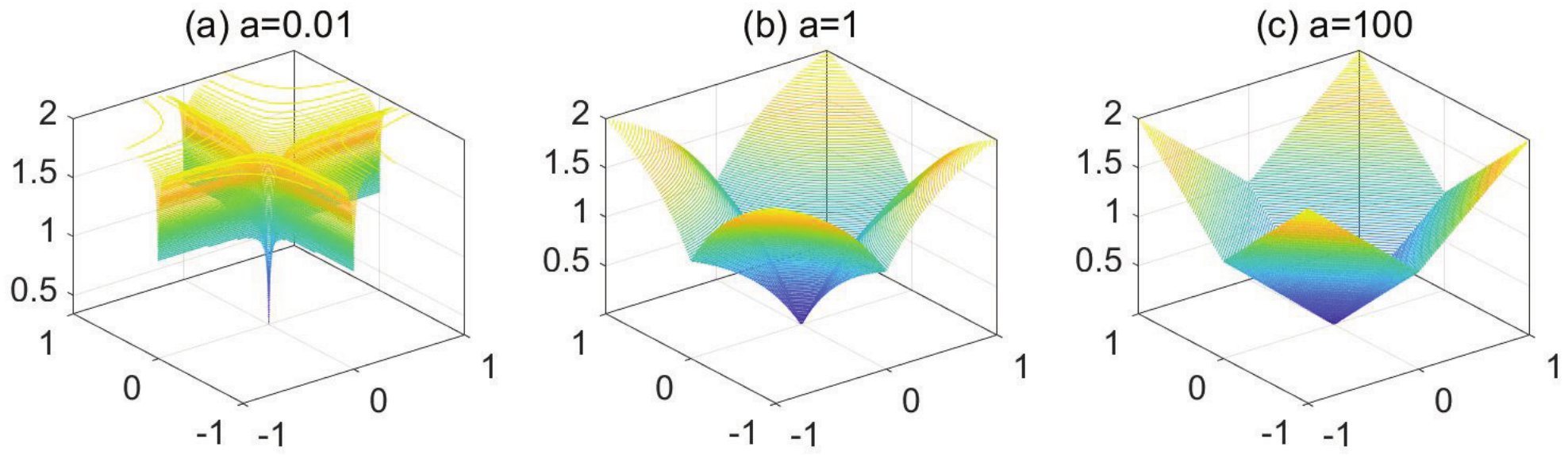}
	\caption{\textbf{Contour plots of T$\ell_1$ with different values of parameter $a$ in two dimensions.} (a) T$\ell_1$ with $a=10^{-2}$. (b) T$\ell_1$ with $a=1$. (c) T$\ell_1$ with $a=10^2$. Compared with the subfigure \ref{Fig.sub.0}, \ref{Fig.sub.10} and \ref{Fig.sub.5}, we can see that the contour plot of T$\ell_1$ with $a=10^{-2}$ is similar to the contour plot of $\ell_0$ while T$\ell_1$ with $a=10^2$ is similar to $\ell_1$. In addition, T$\ell_1$ with $a=1$ looks like $\ell_{1/2}$.}	
	\label{aofTL1}
\end{figure}

When acting on vectors, the definition of T$\ell_1$ can be formulated as
\begin{align}
T\ell_1(\mathbf{x})=\sum_{i=1}^{N}\rho_a(x_i), \  \forall\! \ \mathbf{x}=(x_1,x_2,...,x_N)^T\in \mathbb{R}^N.
\end{align}
With the change of parameter $a$, T$\ell_1$ interpolates $\ell_0$ and $\ell_1$ norm as,
\begin{equation}
\begin{aligned}
\lim_{a \to 0^+}T\ell_1(\mathbf{x})=\sum_{i=1}^{N}I_{\left\{x_i\neq 0\right\}}=\|\mathbf{x}\|_0, \ \lim_{a \to +\infty} T\ell_1(\mathbf{x}) = \sum_{i=1}^N \left|x_i\right|= \|\mathbf{x}\|_1.
\end{aligned}
\end{equation}
As a demonstration, we plot the contours of T$\ell_1$ with $a=10^{-2},1,10^2$ in Fig. \ref{aofTL1}. From the set of figures, we can observe that T$\ell_1$ can indeed approximate $\ell_0, \ell_{1/2}$ and $\ell_1$ well with the adjustment of parameter $a$.

The work \cite{zhang2017transformed} extends the T$\ell_1$ to matrix space based on the singular values in matrix completion problem, which interpolates the rank and the nuclear norm through the nonnegative parameter $a$. In this work, since the sparsity of neural networks is introduced in the component wise for weights of layers, we propose the T$\ell_1$ for matrix in the following form:
\begin{align}
T\ell_1(X) = \sum_{i,j}\rho_a(x_{i,j}),
\end{align}
where $x_{i,j}$ is the element of $i$-th row and $j$-th column in matrix $X\in \mathbb{R}^{m\times n}$. Then, the regularization function $\Omega(W^{(l)})$ in \eqref{objec} becomes $T\ell_1(W^{(l)})$. We choose T$\ell_1$ as the sparse regularizer based on the following reasons. Firstly, compared with the convex regularizers such as $\ell_1$, T$\ell_1$ is unbiased \cite{lv2009unified} and can produce sparser solution~\cite{zhang2017minimization}. Secondly, compared with non-Lipschitz regularizers such as $\ell_p$, the trending rate of T$\ell_1$ can be controlled. Thirdly, compared with the piecewise regularizers such as SCAD and MCP, the formula of T$\ell_1$ is more concise. Last but not least, compared with non-parameter $\ell_{1-2}$, T$\ell_1$, which relies on a parameter $a$, is adjustable for various tasks.

Besides removing as many unnecessary connections as possible, reducing the number of neurons also plays a powerful role in light weight neural networks. Group sparsity, which requires that the elements in one group are all zero, or none of them is, is a typical way of removing neurons and has been employed in several works~\cite{zhou2016less,alvarez2016learning,scardapane2017group,lebedev2016fast}. In this work, to further minify the size of the networks and reduce the computation complexity, we consider using group sparsity as an assist of T$\ell_1$ and propose the integrated transformed $\ell_1$ regularizer in the following way:
\begin{align}\label{IGTL1}
\Omega(W^{(l)})=\mu_l T\ell_1(W^{(l)})+(1-\mu_l)\sum_g \| W^{(l)}_g \|_2,
\end{align}
where $g\in\mathscr{G}$ is the weight group obtained by dividing the weight matrix according to neurons. $W^{(l)}_g$ denotes the weight vector for group $g$ defined on $W^{(l)}$. $\mu_l$ is a positive parameter that balances the T$\ell_1$ term and the group sparsity term. In the above regularizer, group sparsity, which uses the $\ell_1$ norm to zero out variables that are grouped by the $\ell_2$ norm, helps automatically decide the number of neurons in each layer. T$\ell_1$ plays the role of inducing sparsity in the connection level. Both of them work together to decide the suitable number of neurons and promote sparsity among the remaining simultaneously.

Applying the regularization function \eqref{IGTL1} to problem \eqref{objec}, our model of transformed $\ell_1$ regularization for learning sparse deep neural networks can be formulated as:
\begin{align}\label{objec3}
\min_{\{W^{(l)}\}}\mathcal{L}\left(\{W^{(l)}\},\mathcal{T}\right)+\lambda\sum_{l=1}^{L}\left(\mu_l T\ell_1(W^{(l)})+(1-\mu_l)\sum_g \| W^{(l)}_g \|_2\right).
\end{align}
When the training process of \eqref{objec3} terminates, the weights of some connections will turn to zero since they have slight effect on the final performance and then be removed from the network. If all the outgoing or ingoing connections of a neuron are removed, this neuron will be removed as well. Afterwards, a sparse network with less neurons and connections can be yielded. Although there have been several works that used regularization term to promote sparsity in DNNs, to the best of our knowledge, this is the first work to use non-convex regularizer to achieve the same goal. Since non-convex regularizers tend to outperform convex ones in terms of sparsity-promoting effect, our integrated T$\ell_1$ regularizer should be able to obtain network with a sparser structure intuitively.

\subsection{The Optimization Algorithm}
In this subsection, we will focus on how to train the model \eqref{objec3} efficiently. The integrated T$\ell_1$ regularizer is non-smooth, causing increased difficulty in solving \eqref{objec3}. Proximal methods, which are popularly used to handle with non-smooth problems, draw our attention. They can be interpreted as solving optimization problems by finding fixed points of appropriate operators. They are often conceptually and mathematically simple and can work fast under extremely general conditions~\cite{parikh2014proximal}. Therefore, we consider using proximal gradient method to solve the model. 

The proximal gradient approach iteratively minimizes \eqref{objec3} layer by layer through the following formula:
\begin{align}\label{pr}
W_{t+1}^{(l)}=prox_{\lambda\gamma\Omega}(W_t^{(l)}-\gamma \nabla\mathcal{L}(W_t^{(l)},\mathcal{T})),
\end{align}
where $\gamma$ is the step size. $W_t^{(l)}$ represents the variable of $l$-th layer in $t$-th iteration and $W_{t+1}^{(l)}$ is the variable of $l$-th layer obtained after current iteration. The $prox$ denotes proximal operator whose definition on a function $f$ is formulated as
\begin{align}\label{proxf}
prox_f(X)=\mathop{\arg\min}_Y \left\{f(Y)+(1/2)\|Y-X\|_2^2\right\},
\end{align}
where $\|\cdot\|_2$ is the Euclidean norm. Therefore, the \eqref{pr} can be expanded as:
\begin{align}\label{promodel}
W_{t+1}^{(l)}=\mathop{\arg\min}_{W^{(l)}} \left\{\Omega(W^{(l)}) + \frac{1}{2\lambda \gamma}\| W^{(l)}-(W_t^{(l)}-\gamma \nabla\mathcal{L}(W_t^{(l)},\mathcal{T}))\|_2^2\right\}.
\end{align}

In this work, since the problem we solve is in DNNs, it is costly to compute the gradient on whole training dataset for each update. In fact, in DNNs, stochastic gradient descent (SGD) rather than standard gradient descent is the commonly used optimization procedure~\cite{bottou1991stochastic}. Thus, we use SGD to replace the gradient descent step in \eqref{promodel}. In details, SGD involves computing the outputs and errors, calculating the average gradient on a few instances and adjusting the weights accordingly. Then, \eqref{promodel} turns into
\begin{align}\label{promodel2}
W_{t+1}^{(l)}\!=\!\mathop{\arg\min}_{W^{(l)}}\!\! \left\{\!\!\Omega(W^{(l)}\!) \!\!+\!\! \frac{1}{2\lambda \gamma}\| W^{(l)}\!\!-\!\!(W_t^{(l)}\!\!-\!\!\gamma\! \sum_{i=1}^{n}\nabla\mathcal{L}(W_t^{(l)}\!,\!\{x_i,y_i\})/n)\|_2^2\!\!\right\}\!\!,
\end{align}
where $n$ is the mini-batch size in SGD and $\{x_i,y_i\}_{i=1}^{n}$ are the $n$ samples randomly selected from the dataset $\mathcal{T}$.

The regularization term $\Omega(W^{(l)})$ in our objective function is combined by two single regularizer and computing the proximal operator of such regularizer is not easy. Fortunately, the proximal gradient method can avoid this procedure and only proximal operator of each single regularizer is required. To be more specific, the objective function \eqref{objec3} can be solved simply by iteratively implementing update on the variables layer by layer through the proximal operators of two regularizers in succession after performing a gradient step on the variables based on the loss term, i.e.
\begin{align}\label{obpr}
W_{t+1}^{(l)}\!=\!prox_{\lambda\gamma(1-\mu_l)GS}(prox_{\lambda\gamma\mu_lT\ell_1}(W_t^{(l)}-\gamma \sum_{i=1}^{n}\nabla\mathcal{L}(W_t^{(l)}\!,\!\{x_i,y_i\})/n)).
\end{align}

To calculate \eqref{obpr}, the proximal operators of T$\ell_1$ and group sparsity are required. As can be seen from \eqref{proxf}, computing the proximal operator of a convex function turns into solving a small convex regularized optimization problem, which usually obtains a closed-form solution, for example, group sparsity in our model. The proximal operator of group sparsity is formulated as:
\begin{align}\label{proxGS}
prox_{\lambda\gamma(1-\mu_l) GS}(W^{(l)})= \left(1-\lambda\gamma(1-\mu_l)/\vert\vert \mathbf{w}_g^{(l)}\vert\vert_2\right)_+ w_{g,i}^{(l)},
\end{align}
for all $g$ and $i$, where $g$ is a group, and $i$ is the index in each group. For example, $w^{(p)}_{m,n}$ represents the $n$-th element in $m$-th group of $p$-th layer. However, for some non-convex functions, their proximal operators might not have closed forms in general, like $\ell_p$ penalty with $p\in(0,1)$. Next, we will prove that there indeed exists closed-formed expression for the proximal operator of T$\ell_1$ even though it is non-convex. As mentioned earlier, the proximal operator of T$\ell_1$ can be obtained by solving the optimization problem as follows,
\begin{align}\label{TL1}
\min_{\hat{W}^{(l)}} \ \ \frac{1}{2\lambda\gamma\mu_l}\| \hat{W}^{(l)}-W^{(l)} \|^2_2+T\ell_1(\hat{W}^{(l)}).
\end{align}	
Expanding the optimization problem above yields:
\begin{align}\label{TL12}
\min_{\hat{W}^{(l)}}\ \ \sum_i\sum_j \left(\frac{1}{2\lambda\gamma\mu_l}(\hat{w}^{(l)}_{i,j}-w^{(l)}_{i,j})^2+ \frac{(a+1)\vert \hat{w}^{(l)}_{i,j}\vert}{a+\vert \hat{w}^{(l)}_{i,j}\vert}\right).
\end{align}	
Thus, \eqref{TL1} can be optimized for each $i$ and $j$ respectively, i.e., it can be solved by optimizing
\begin{align}\label{TL1_each}
\min_{\hat{w}^{(l)}_{i,j}} \ \  \frac{1}{2\lambda\gamma\mu_l}(\hat{w}^{(l)}_{i,j}-w^{(l)}_{i,j})^2+ \frac{(a+1)\vert \hat{w}^{(l)}_{i,j}\vert}{a+\vert \hat{w}^{(l)}_{i,j}\vert}
\end{align}
for each $i$ and $j$. This is an unconstrained optimization problem with univariable $\hat{w}^{(l)}_{i,j}$, whose solution can be obtained by calculating its subgradient. The optimal solution of \eqref{TL1_each} is formulated as follows,
\begin{eqnarray}
\hat{w}_{i,j}^{(l)}=
\left\{
\begin{array}{lll}
0, \ \ &if \ \ \vert w^{(l)}_{i,j}\vert\leq t \\
g_{\lambda\gamma\mu_l} (w^{(l)}_{i,j}), \ \ & otherwise
\end{array}
\right.
\end{eqnarray}
for all $i$ and $j$, where $g_{\lambda\gamma\mu_l}(w)$ is defined as follows,
\begin{align}\label{g}
g_{\lambda\gamma\mu_l}(w)=sgn(w)\left\{2(a\!+\!\vert w \vert)\cos ({\varphi(w)}/3)/3-2a/3+\vert w\vert/3\right\}
\end{align}
with $\varphi(w)=\arccos\left(1-\frac{27\lambda\gamma\mu_l a(a+1)}{2(a+\vert w\vert)^3}\right)$, and $t$ is given as follows,
\begin{eqnarray}\label{t}
t=
\left\{
\begin{array}{lll}
\lambda\gamma\mu_l (a+1)/a, \ \ &if \ \ \lambda\gamma\mu_l\leq \frac{a^2}{2(a+1)} \\
\sqrt{2\lambda\gamma\mu_l (a+1)}-a/2, \ \ &otherwise.
\end{array}
\right.
\end{eqnarray}
More details for the solving process can be found in~\cite{zhang2017minimization}. Therefore, the proximal operator of T$\ell_1$ can be formulated as
\begin{eqnarray}\label{proxTL1}
prox_{\lambda\gamma\mu_l T\ell_1}(W^{(l)})=
\left\{
\begin{array}{lll}
0, \ \ &if \ \ \vert w^{(l)}_{i,j}\vert\leq t \\
g_{\lambda\gamma\mu_l} (w_{i,j}^{(l)}), \ \ & otherwise
\end{array}
\right.
\end{eqnarray}
for each $i$ and $j$, in which $g_{\lambda\gamma\mu_l}$ and $t$ is defined as \eqref{g} and \eqref{t}, respectively.

We summarize the whole optimization process in Algorithm~\ref{pro}. In the algorithm, the stopping criterion is predefined and the common used one is the increase of loss between two consecutive steps is less than a threshold or the maximum number of iterations is achieved. 

\begin{algorithm}
	\caption{Stochastic Proximal Gradient Descent for Model \eqref{objec3}}
	\label{pro}
	\hspace*{0.02in} {\bf Input:}
	initial weight matrix $W_0$, regularization parameter $\lambda$, balancing parameter for each layer $\mu_l$, learning rate $\gamma$, mini-batch size $n$, training dataset $\mathcal{T}$
	\begin{algorithmic}
		\STATE $t=1$
		\REPEAT
        \STATE Randomly select $n$ samples from $\mathcal{T}$
        \FOR{each layer $l$}
        \FOR{each sample $\{x_i,y_i\}$ in the $n$ samples selected}
        \STATE $L_i^{(l)}:=\nabla\mathcal{L}(W_{t-1}^{(l)},\{x_i,y_i\})$
        \ENDFOR
        \STATE $L^{(l)}:=\sum_{i=1}^{n}L_i^{(l)}/n$
		\STATE $W^{(l)}_t :=W^{(l)}_{t-1}-\gamma L^{(l)}$
		\STATE Update $W^{(l)}_t$ by \eqref{proxTL1}: $W^{(l)}_t :=prox_{\lambda \gamma\mu_l T\ell_1}(W^{(l)}_t)$
		\STATE Update $W^{(l)}_t$ by \eqref{proxGS}: $W^{(l)}_t :=prox_{\lambda \gamma(1-\mu_l) GS}(W^{(l)}_t)$
		\ENDFOR
		\STATE $t :=t+1$
		\UNTIL{some stopping criterion is satisfied}
	\end{algorithmic}
	\hspace*{0.02in} {\bf Output:}
	The solution $W_{t-1}$
\end{algorithm}

\section{Experiments}
In this section, we evaluate the proposed combined regularizer on several real-world datasets. The regularizer is applied to all layers of the network, except the bias term.
\subsection{Baselines and Datasets}
To demonstrate the superiority of the integrated T$\ell_1$, we compare it with several state-of-the-art baselines:
\begin{itemize}
	\item $\bm{\ell_1}$. Network regularized by $\ell_1$, which produces global sparsity.
	\item \textbf{Sparse Group Lasso (SGL)}. Network with a regularizer that combines group sparsity and $\ell_1$ regularizer~\cite{scardapane2017group}.
	\item \textbf{Combined Group and Exclusive Sparsity (CGES)}. Network with a regularizer that combines group sparsity and exclusive sparsity~\cite{yoon2017combined}.
\end{itemize}

We select six public classification datasets which are commonly used in DNNs to conduct experiments,
\begin{itemize}
	\item \textbf{DIGITS}. This is a toy dataset of handwritten digits, composed of 1,797 $8\times 8$ grayscale images. We use this dataset to illustrate the effect of parameter $a$ in the integrated T$\ell_1$ regularizer and the sparsity-promoting capacity of several regularizers. 
	\item \textbf{MNIST}. This dataset consists of 70,000 $28\times28$ grayscale images of handwritten digits, which can be classified into 10 classes. The number of training instances and test samples are 60,000 and 10,000, respectively. 
	\item \textbf{Fashion-MNIST}. This dataset consists of a training set with 60,000 instances and a test set with 10,000 examples. Each example is a $28\times28$ grayscale image, associated with a label from 10 classes (T-shirt, trouser, pullover, dress, coat, sandal, shirt, sneaker, bag and ankle boot). Fashion-MNIST serves as a direct drop-in replacement for the original MNIST dataset. 
	\item \textbf{PENDIGITS}. This dataset is composed of 10,992 $4\times4$ grayscale images of handwritten digits 0-9, where there is 7,494 training instances and 3,498 test samples. 
	\item \textbf{Sensorless Drive Diagnosis (SDD)}. This dataset is downloaded from the UCI repository. It contains 58,508 examples obtained under 11 different operating conditions. In this dataset, we need to predict a motor with one or more defective components, starting from a set of 48 features obtained from electric drive signals of the motor. 
	\item \textbf{CIFAR-10}. This dataset consists of 60,000 $32\times 32$ colour images in 10 classes (airplane, automobile, bird, cat, deer, dog, frog, horse, ship and truck), with 6,000 images per class. The dataset is divided into one test batch with 10,000 images and five training batches with 10,000 instances in each batch.
\end{itemize}

\subsection{Experimental Setup}
We use Tensorflow framework to implement the models. In all cases, we employ the ReLU function $f(x)=\max(0,x)$ as the activation function. As for the output layer, we apply the softmax activation function. If $\mathbf{x}\in\mathbb{R}^n$ denotes the value that is input to softmax, the $i_{th}$ output can be obtained as,
\begin{align}
f(x_i)=\frac{exp(x_i)}{\sum_{j=1}^{n}exp(x_j)}.
\end{align}
Besides, one-hot encoding is used to encode different classes. We initialize the weights of the network by Xavier or random initialization according to a normal distribution. Networks of MNIST, Fashion-MNIST and PENDIGITS are trained with stochastic gradient descent algorithm, while the rest datasets utilize the popular Adam algorithm. The size of mini-batch is varied depending on the scale of the datasets. We choose the standard cross-entropy loss as the loss function. In the experiments, we would like group sparsity to play the leading role in the lower layers while T$\ell_1$ regularizer has more effect at the top layers, just as mentioned in \cite{yoon2017combined}. To this end, we dynamically set $\mu_l = s + (1-2s)(l-1)/(L-1)$, where $L$ is the number of layers, $l\in \{1,2,...,L\}$ is the index of each layer and $s$ is the lowest value that can be used for the T$\ell_1$ term. The regularization parameter $\lambda$ and the parameter $a$ in T$\ell_1$ are selected through the grid search technique, with $\lambda$ varying from $10^{-6}$ to $10^{-4}$ and $a$ in $\{10^{-3},10^{-2}, 10^{-1}, 1, 10, 10^2\}$. On one specific dataset, we use the same network architecture for various penalties to keep the comparison fair. The detailed network architecture setting for each dataset is presented in Table \ref{archi}, where the second column and the third column denote the number of convolutional layers and fully connected layers, respectively. To obtain more reliable results, we repeatedly run the training process three times. The final results are reported as an average with standard deviations over these three times.

\begin{table*}[!t]
	\centering\small	
	\caption{Network architecture for each dataset}
	\label{archi}{
		\begin{tabular}{ccc}
			\toprule
			Dataset& $\#$ convolutional layers &  $\#$ fully connected layers\\
			\midrule
			DIGITS & $1$ & $2$\\
			\midrule
			MNIST & $2$ & $2$\\
			\midrule
			Fashion-MNIST & $2$ & $3$\\
			\midrule
            PENDIGITS & $2$ & $2$\\
			\midrule
            SDD & $2$ & $3$\\
			\midrule
            CIFAR-10 & $4$ & $2$\\
			\bottomrule
		\end{tabular}
	}
\end{table*}

\begin{table}
	\setlength{\tabcolsep}{2.5pt}
	\centering
	\scriptsize	
	\caption{Performance of different methods on various datasets.}The best results are highlighted in bold face.
	\label{performance}{
	\begin{tabular}{cccccc}
		\toprule
		Dataset& Measure&$\ell_1$ & SGL &CGES & integrated T$\ell_1$\\
		\midrule
		\multirow{3}*{MNIST} & accuracy &$0.9749$$\pm$$0.0018$&$\mathbf{0.9882}$$\mathbf{\pm}$$\mathbf{0.0007}$&$0.9769$$\pm$$0.0008$&$0.9732$$\pm$$0.0007$\\
		~&FLOP&$0.6859$$\pm$$0.0215$&$0.8134$$\pm$$0.0134$&$0.6633$$\pm$$0.0181$&$\mathbf{0.1742}$$\mathbf{\pm}$$\mathbf{0.0006}$\\
		~&parameter&$0.2851$$\pm$$0.0252$&$0.4982$$\pm$$0.0361$&$0.2032$$\pm$$0.0295$&$\mathbf{0.1601}$$\mathbf{\pm}$$\mathbf{0.0029}$\\
		\midrule
		\multirow{3}*{Fashion-MNIST} & accuracy &$\mathbf{0.8947}$$\mathbf{\pm}$$\mathbf{0.0009}$&$0.8927$$\pm$$0.0006$&$0.8804$$\pm$$0.0233$&$0.8873$$\pm$$0.0026$\\
		~&FLOP&$0.7056$$\pm$$0.0034$&$0.7038$$\pm$$0.0015$&$0.7446$$\pm$$0.0021$&$\mathbf{0.3097}$$\mathbf{\pm}$$\mathbf{0.0099}$\\
		~&parameter&$0.2323$$\pm$$0.0092$&$\mathbf{0.2275}$$\mathbf{\pm}$$\mathbf{0.0039}$&$0.3362$$\pm$$0.0043$&$0.3102\pm$$0.0076$\\
		\midrule
		\multirow{3}*{SDD} & accuracy &$0.9833$$\pm$$0.0013$& &$\mathbf{0.9909}$$\mathbf{\pm}$$\mathbf{0.0034}$&$0.9897$$\pm$$0.0011$\\
		~&FLOP&$0.4085$$\pm$$0.0429$&--&$0.3072$$\pm$$0.0619$&$\mathbf{0.1584}$$\mathbf{\pm}$$\mathbf{0.0494}$\\
		~&parameter&$0.4053$$\pm$$0.0432$& &$0.3035$$\pm$$0.0622$&$\mathbf{0.1608}$$\mathbf{\pm}$$\mathbf{0.0480}$\\
		\midrule
		\multirow{3}*{PENDIGITS} & accuracy & $0.9715$$\pm$$0.0014$&$0.9739$$\pm$$0.0021$&$0.9732$$\pm$$0.0015$&$\mathbf{0.9745}$$\mathbf{\pm}$$\mathbf{0.0013}$\\
		~&FLOP&$0.4414$$\pm$$0.0265$&$0.7220$$\pm$$0.0038$&$0.6027$$\pm$$0.0075$&$\mathbf{0.3301}$$\mathbf{\pm}$$\mathbf{0.0089}$\\
		~&parameter&$0.4296$$\pm$$0.0241$&$0.7241$$\pm$$0.0034$&$0.5948$$\pm$$0.0114$&$\mathbf{0.3181}$$\mathbf{\pm}$$\mathbf{0.0080}$\\
		\midrule
		\multirow{3}*{CIFAR-10} & accuracy & $0.7716$$\pm$$0.0052$&$0.7667$$\pm$$0.0006$&$0.7775$$\pm$$0.0018$&$\mathbf{0.7797}$$\mathbf{\pm}$$\mathbf{0.0052}$\\
		~&FLOP&$0.8827$$\pm$$0.0007$&$0.6817$$\pm$$0.0076$&$0.8210$$\pm$$0.0180$&$\mathbf{0.2928}$$\mathbf{\pm}$$\mathbf{0.0121}$\\
		~&parameter&$0.5104$$\pm$$0.0041$&$0.8076$$\pm$$0.0127$&$0.7310$$\pm$$0.0035$&$\mathbf{0.2655}$$\mathbf{\pm}$$\mathbf{0.0076}$\\
		\midrule
		\multirow{3}*{Average rank} & accuracy &$3$&$2.4$&$2.4$&$\mathbf{2.2}$\\
		~&FLOP&$3$&$3.2$&$2.8$&$\mathbf{1}$\\
		~&parameter&$2.4$&$3.4$&$2.8$&$\mathbf{1.4}$\\
		\bottomrule
	\end{tabular}
   }
\end{table}

\begin{figure}
	\setlength{\abovecaptionskip}{-0.5cm}
	\centering
	\includegraphics[width=1.05\textwidth]{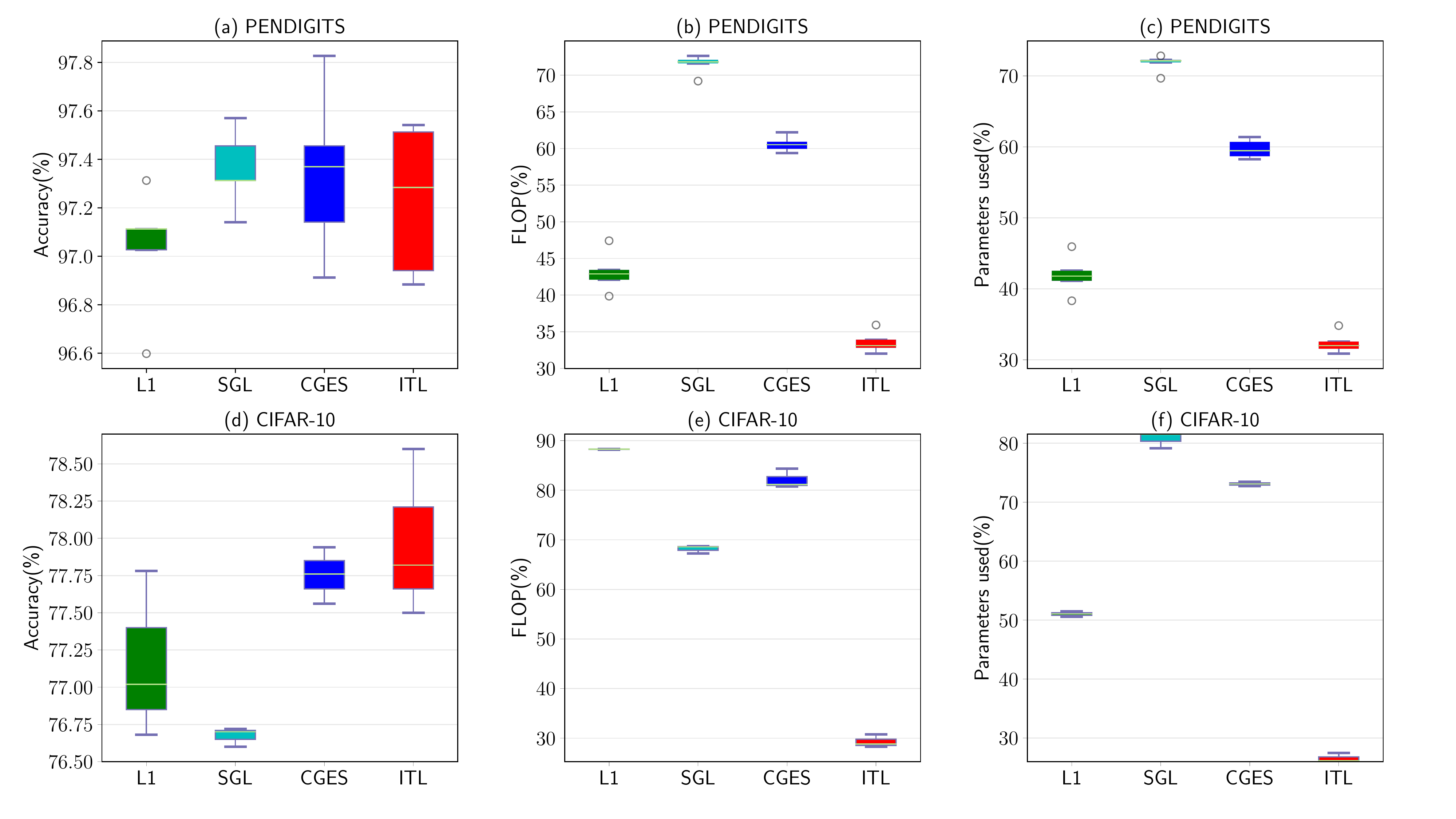}
	\caption{\textbf{Comparisons of classification and sparsity-promoting effect of four algorithms on PENDIGITS and CIFAR-10.} The $x$-axis presents four algorithms: L1 ($\ell_1$), SGL (sparse group lasso in \cite{scardapane2017group}), CGES (combined group and exclusive sparsity in \cite{yoon2017combined}), ITL (integrated transformed $\ell_1$). The $y$-axis is prediction accuracy, FLOP or parameters used in the network. The notched boxes have lines at the lower quartile, median, and upper quartile values. The whiskers are lines extending from each end of the box to the most extreme data value with 1.5$\cdot$IQR (interquartile range) of the box. Outliers, whose value is beyond the ends of the whiskers, are displayed by dots.}
	\label{box} 
\end{figure}

\subsection{Performance of integrated T$\ell_1$ regularizer}
In this subsection, we compare our integrated T$\ell_1$ with several baselines to verify the superiority of our model. To quantitatively measure the performance of various models, three metrics are utilized, including the prediction accuracy, the corresponding number of parameters used in the network and the corresponding number of floating point operations (FLOP). A higher accuracy means that the model can train a better network to implement classification tasks. A lower FLOP indicates that the network can reduce the computation complexity more significantly. The ability of saving memory is reflected by the parameters used in the network. Therefore, the smaller the number of parameters used is, the better the regularizer is.

We list the results in Table \ref{performance}. The average ranks of each method for all three measurements are reported in the last three rows of Table \ref{performance}. The best results are highlighted in bold face. We also present the results on PENDIGITS and CIFAR-10 by boxplots in Fig. \ref{box}. As seen from Table~\ref{performance}, the performance of our model is comparable when compared with other baselines. Our model even achieves the best results in terms of all three indicators on PENDIGITS and CIFAR-10. Furthermore, the average rank of our integrated T$\ell_1$ for each measurement is also the best. Although the prediction accuracy on MNIST obtained by the integrated T$\ell_1$ regularizer is slightly lower than other penalties, the largest gap is $0.018$, which is less significant when compared with the memory and the computation saved. The rate of FLOP for our integrated T$\ell_1$ is $0.1742$, nearly $0.49$ less than the least FLOP of other three competitors. $49\%$ less FLOP means that a large number of computation will be saved. For the parameters used, there is also a reduction of $4\%$ for integrated T$\ell_1$ when compared with the least one among other baselines, $16.01\%$ and $20.32\%$, respectively, resulting in a much sparser network architecture. As for the SDD dataset, the accuracy of our integrated T$\ell_1$ is also slightly lower than that of CGES, which is the highest among the baselines. However, like the performance on MNIST, both the FLOP and parameters used of the integrated T$\ell_1$ have evident decline.

\begin{figure*}
	\centering
	\subfigure[MNIST, Accuracy/FLOP]{
		\label{mnistflop}
		\includegraphics[width=0.45\textwidth]{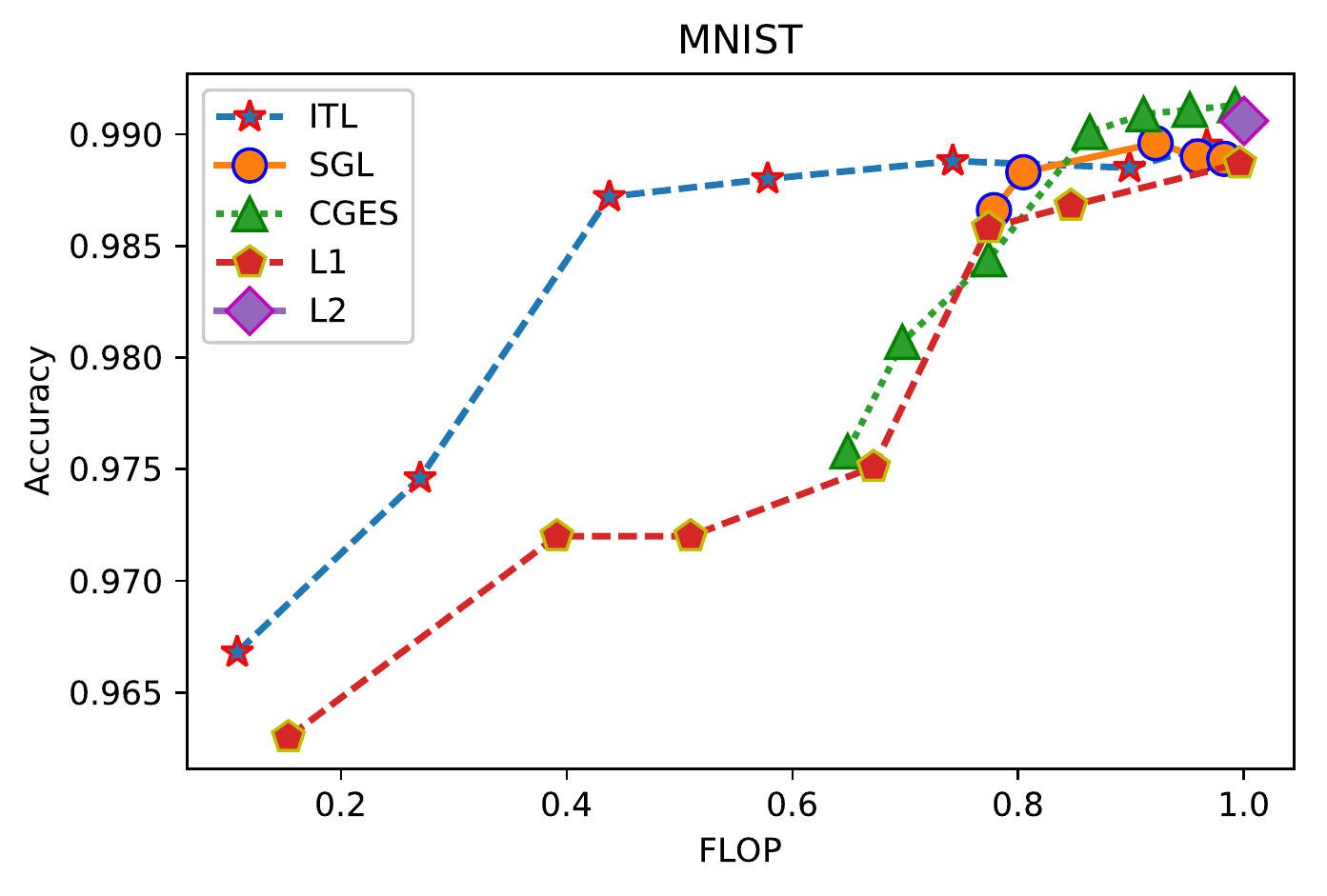}}
	\subfigure[MNIST, Accuracy/Parameters]{
		\label{mnistparam}
		\includegraphics[width=0.45\textwidth]{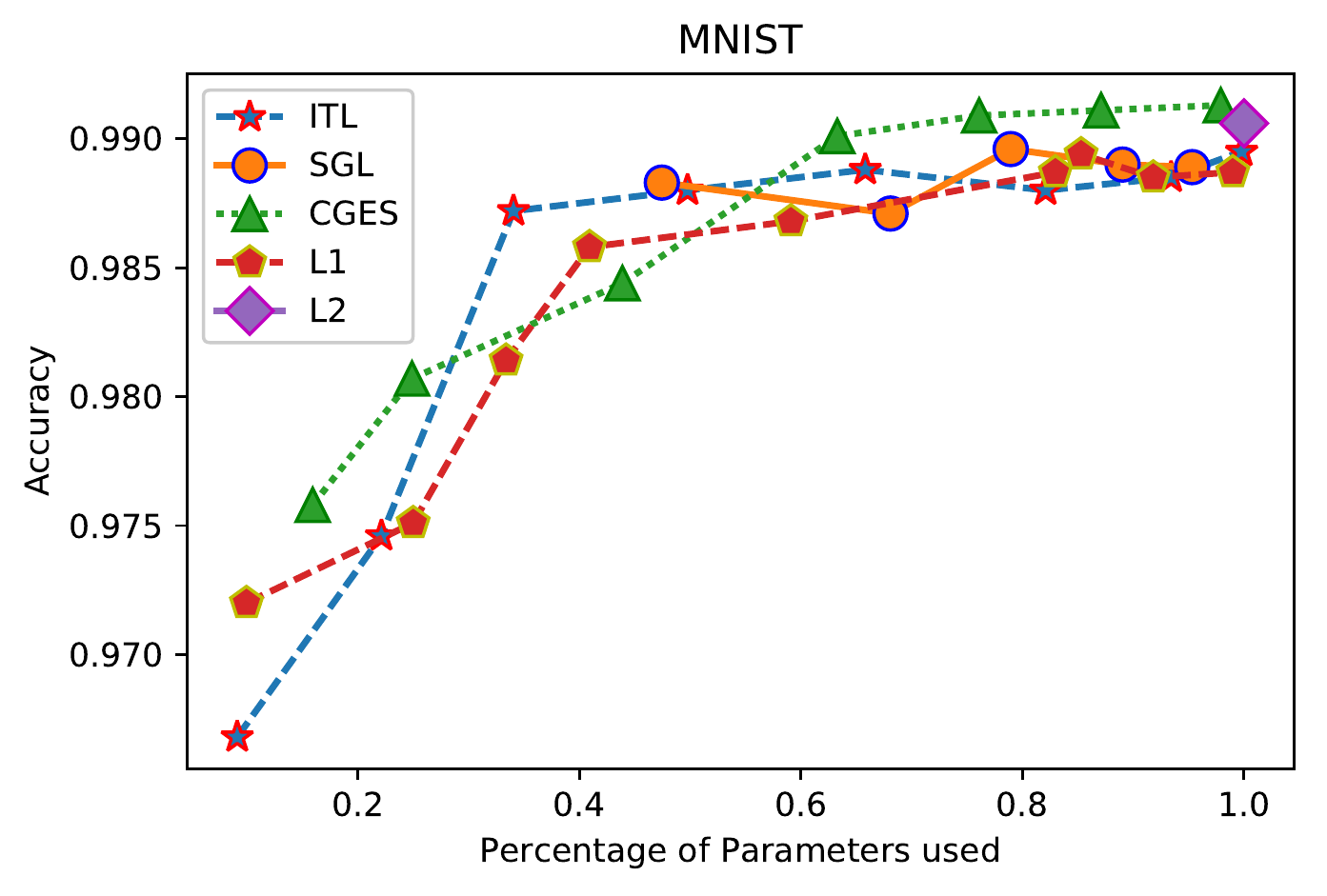}}
	\subfigure[PENDIGITS, Accuracy/FLOP]{
		\label{pendigitsflop}
		\includegraphics[width=0.45\textwidth]{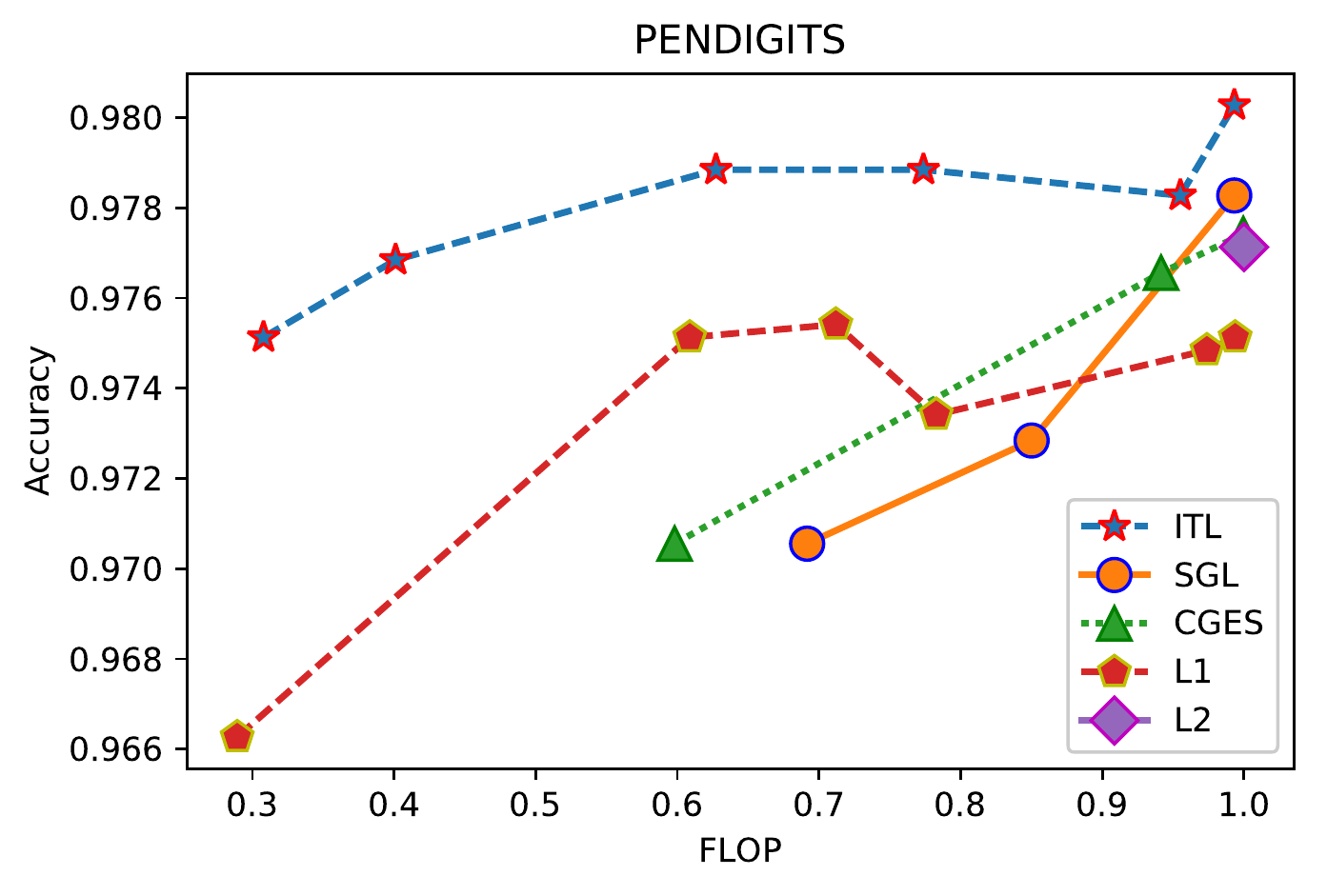}}
	\subfigure[PENDIGITS, Accuracy/Parameters]{
		\label{pendigitspara}
		\includegraphics[width=0.45\textwidth]{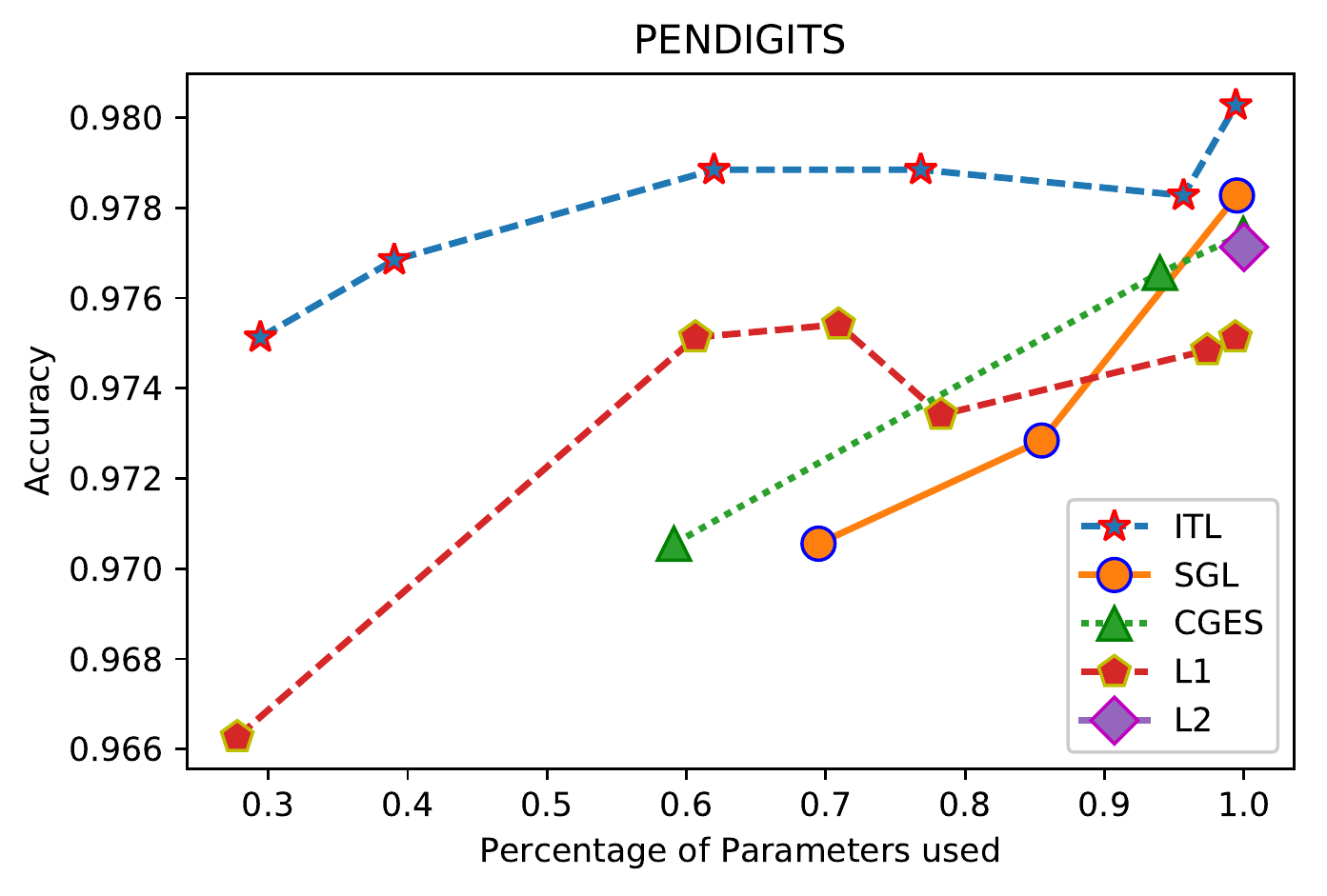}}
	\caption{\textbf{Accuracy-efficiency trade-off.} In order to explore how each regularizer affects the model accuracy at various sparsity levels, we report the accuracy over FLOP and the accuracy over the percentage of parameters used. We obtain the results by varying the regularization parameters. L1 denotes the network with the $\ell_1$ regularizer and L2 denotes the network with the $\ell_2$ regularizer. SGL is the sparse group lasso proposed in the work \cite{scardapane2017group}. CGES is combined group and exclusive sparsity in the work \cite{yoon2017combined}. ITL is our integrated T$\ell_1$.}
	\label{tradeoff}
\end{figure*}

Next, we discuss how the sparsity-inducing regularizer affects the model accuracy. We change the value of regularization parameter to achieve different sparse levels. Networks with $\ell_2$, $\ell_1$, SGL, CGES and our integrated T$\ell_1$ are considered. In this experiment, we adopt two datasets, i.e. MNIST and PENDIGITS. Results are shown in Fig.~\ref{tradeoff}. As seen from Fig. \ref{mnistflop}, the $\ell_1$ regularizer largely reduces FLOP with a slight drop in accuracy for MNIST dataset. There is no significant decrease in FLOP for the networks with SGL and CGES, about $78\%$ and $63\%$ at least, respectively. However, the integrated T$\ell_1$ can significantly reduce the FLOP without much drop of accuracy. When we turn to Fig. \ref{mnistparam}, sparsity of parameters adopted in the networks regularized by these penalties have little significant difference on the final prediction accuracy. In Fig. \ref{pendigitsflop} and Fig. \ref{pendigitspara}, we can observe that the performance of integrated T$\ell_1$ is obviously better than other regularizers on PENDIGITS. The FLOP and parameters used in the network regularized by integrated T$\ell_1$ can achieve a quite sparse level with a similar performance. CGES and SGL cannot obtain comparable accuracy when the network is equipped with less than $60\%$ of parameters and $60\%$ of FLOP. Although both the integrated T$\ell_1$ and $\ell_1$ regularized networks can obtain comparable accuracy when the parameters of network are only $30\%$, the prediction accuracy of our integrated T$\ell_1$ is much better than that of $\ell_1$, to be more specific, $0.975$ and $0.966$, respectively.

\begin{figure}
	\setlength{\abovecaptionskip}{-0.1cm}
	\centering
	\includegraphics[width=0.7\textwidth]{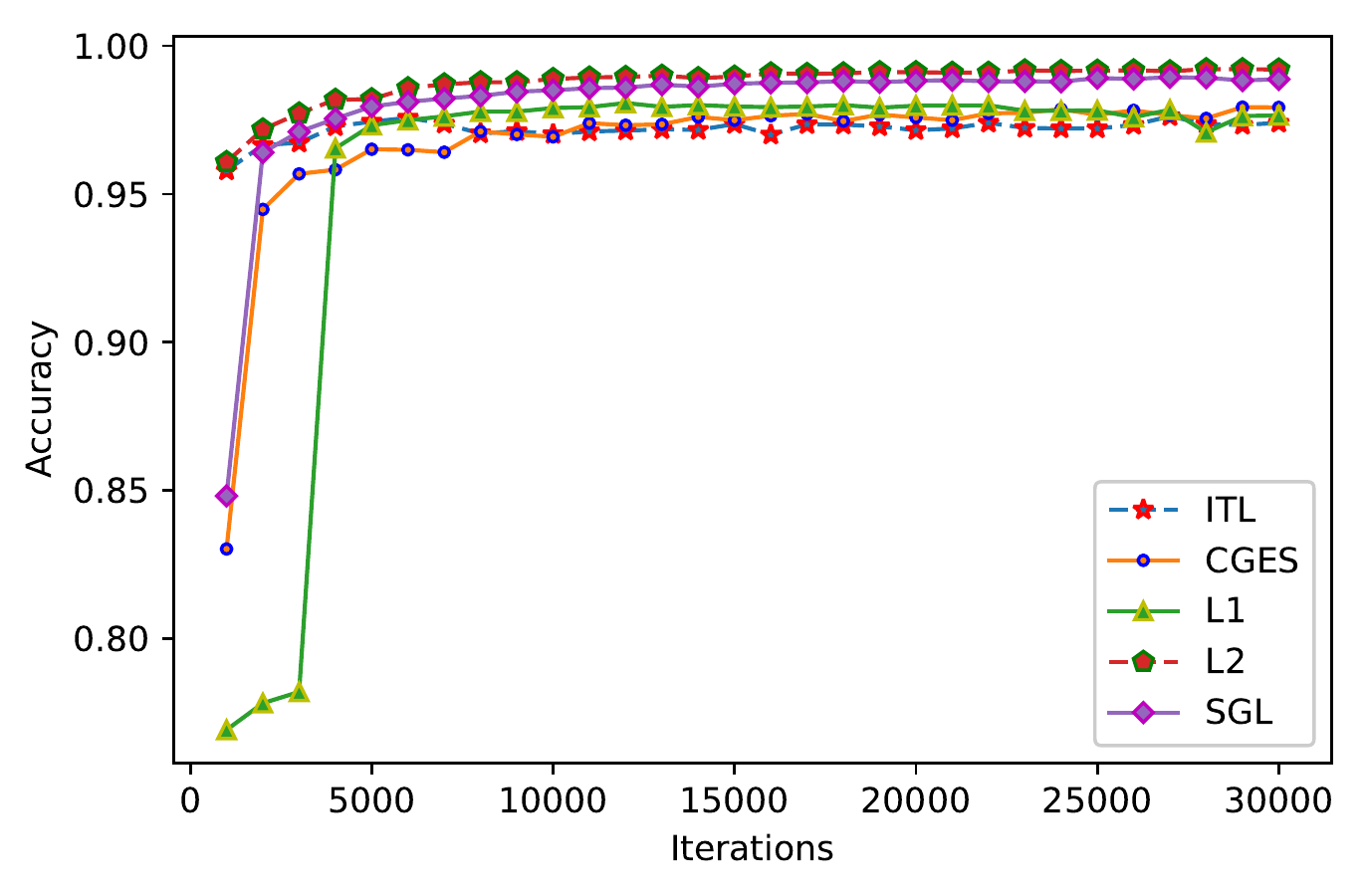}
	\caption{\textbf{Convergence speed on MNIST.} Convergence of networks regularized by $\ell_2$ (denoted by L2), $\ell_1$ (denoted by L1), SGL, CGES and integrated T$\ell_1$ (denoted by ITL).}
	\label{convergencespeed}
\end{figure}

\begin{figure*}
	\centering
	\subfigure[$\ell_1$, Sparsity: $0.00\%$]{
		\label{l1sparsity}
		\includegraphics[width=0.4\textwidth]{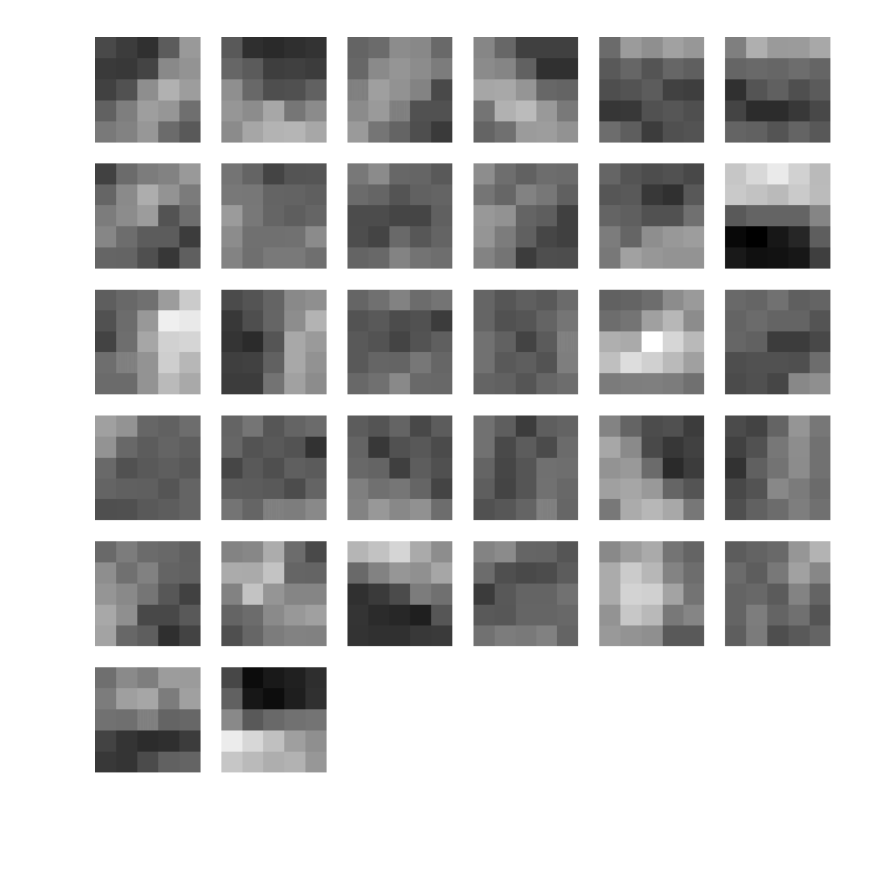}}
	\subfigure[SGL, Sparsity: $0.00\%$]{
		\label{SGLsprasity}
		\includegraphics[width=0.4\textwidth]{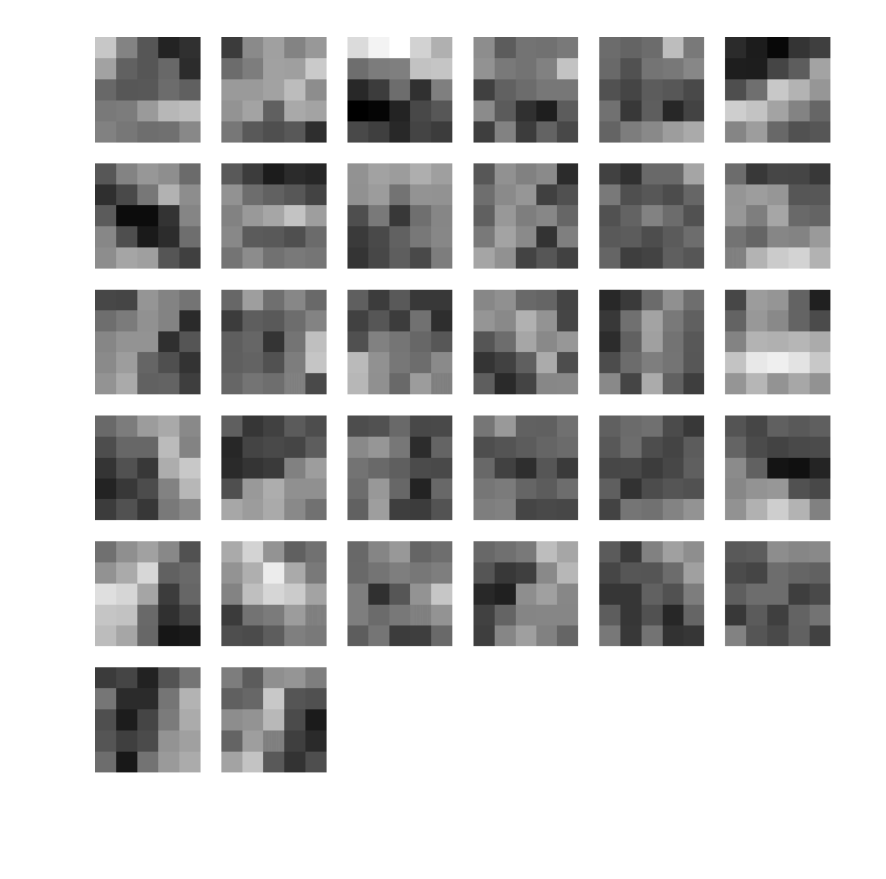}}
	\subfigure[CGES, Sparsity: $1.12\%$]{
		\label{cgessparsity}
		\includegraphics[width=0.4\textwidth]{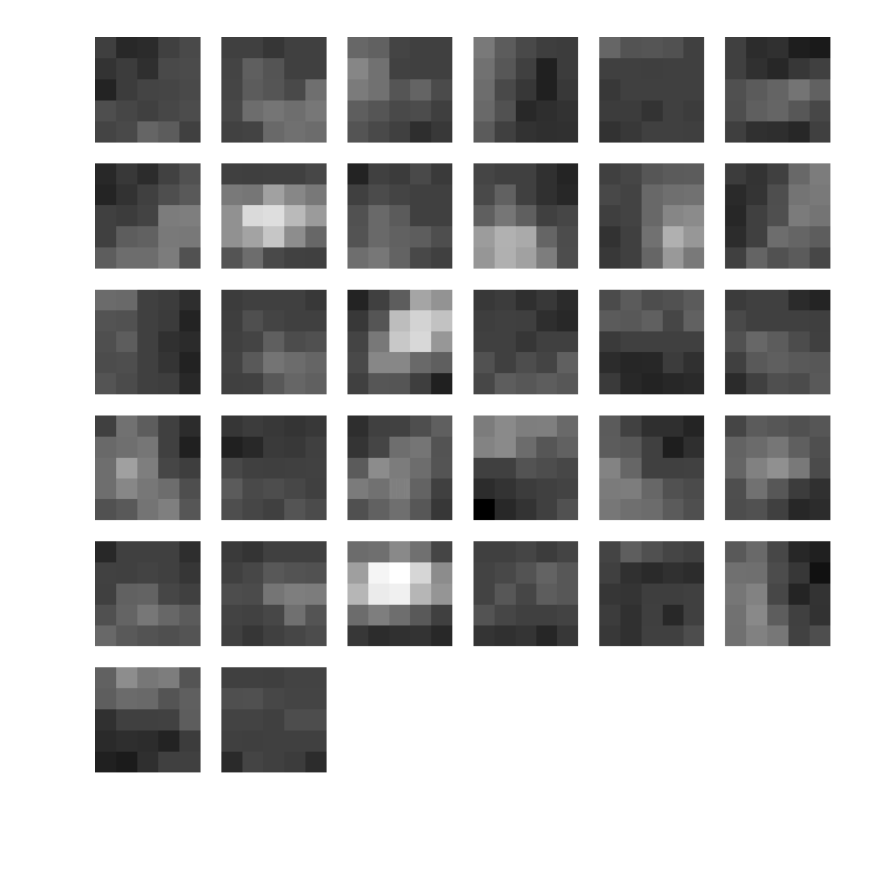}}
	\subfigure[integrated T$\ell_1$, Sparsity: $44.50\%$]{
		\label{igtlsparsity}
		\includegraphics[width=0.4\textwidth]{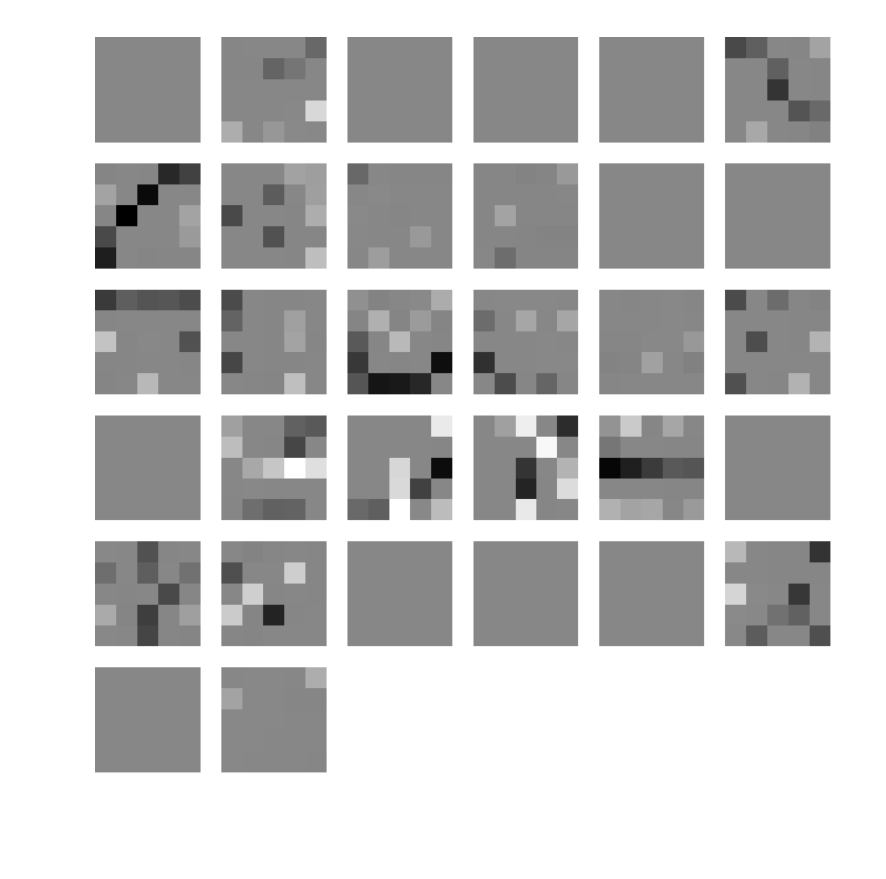}}
	\caption{\textbf{Visualization of filters of the first convolutional layer for the network trained on MNIST.}  The $\ell_1$ regularizer and SGL regularizer result in smooth non-sparse filters, while CGES obtains filters with a slight sparse level. In contrast, integrated T$\ell_1$ completely removes redundant features and obtains much shaper filters.}
	\label{visualization}
\end{figure*}

In real-world applications, the network, which converges with less iterations, is more desirable. Therefore, in this experiment, we discuss the empirical convergence speed of the regularizers on MNIST. In Fig. \ref{convergencespeed}, it is obvious that the network with integrated T$\ell_1$ achieves a comparable accuracy with fewer iteration steps than $\ell_1$, SGL and CGES. In details, when the number of iterations is about $1000$, the network with our integrated T$\ell_1$ can achieve an accuracy of nearly $0.95$ while the accuracy of $\ell_1$, SGL and CGES is less than $0.85$.  

Next, we visualize the sparsity of filters in the first convolutional layer for the network trained on MNIST and display the results in Fig. \ref{visualization}. In our network architecture, $\ell_1$ regularizer and SGL have little effect on the sparsity of the filters. The CGES results in a little sparsity. In contrast, our integrated T$\ell_1$ zeroes out some spatial features when they compete with other filters, resulting in much shaper filters, compared with other regularizers. Therefore, there is less redundance among filters as the network trained with other regularizers.

\begin{figure*}[!htbp]
	\centering
	\subfigure[Accuracy]{
		\label{accuracy}
		\includegraphics[width=0.48\textwidth]{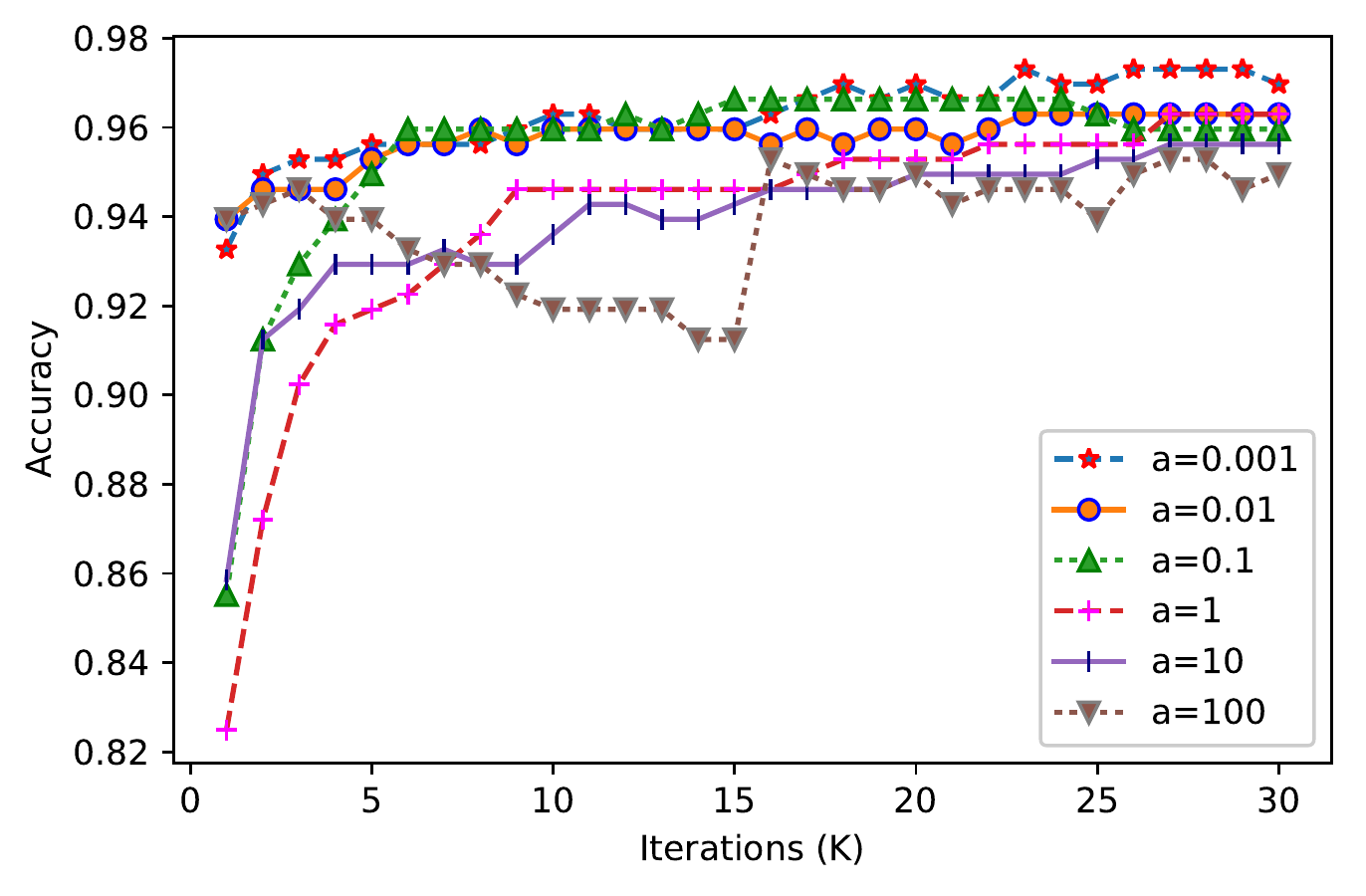}}
	\subfigure[FLOP]{
		\label{flop}
		\includegraphics[width=0.48\textwidth]{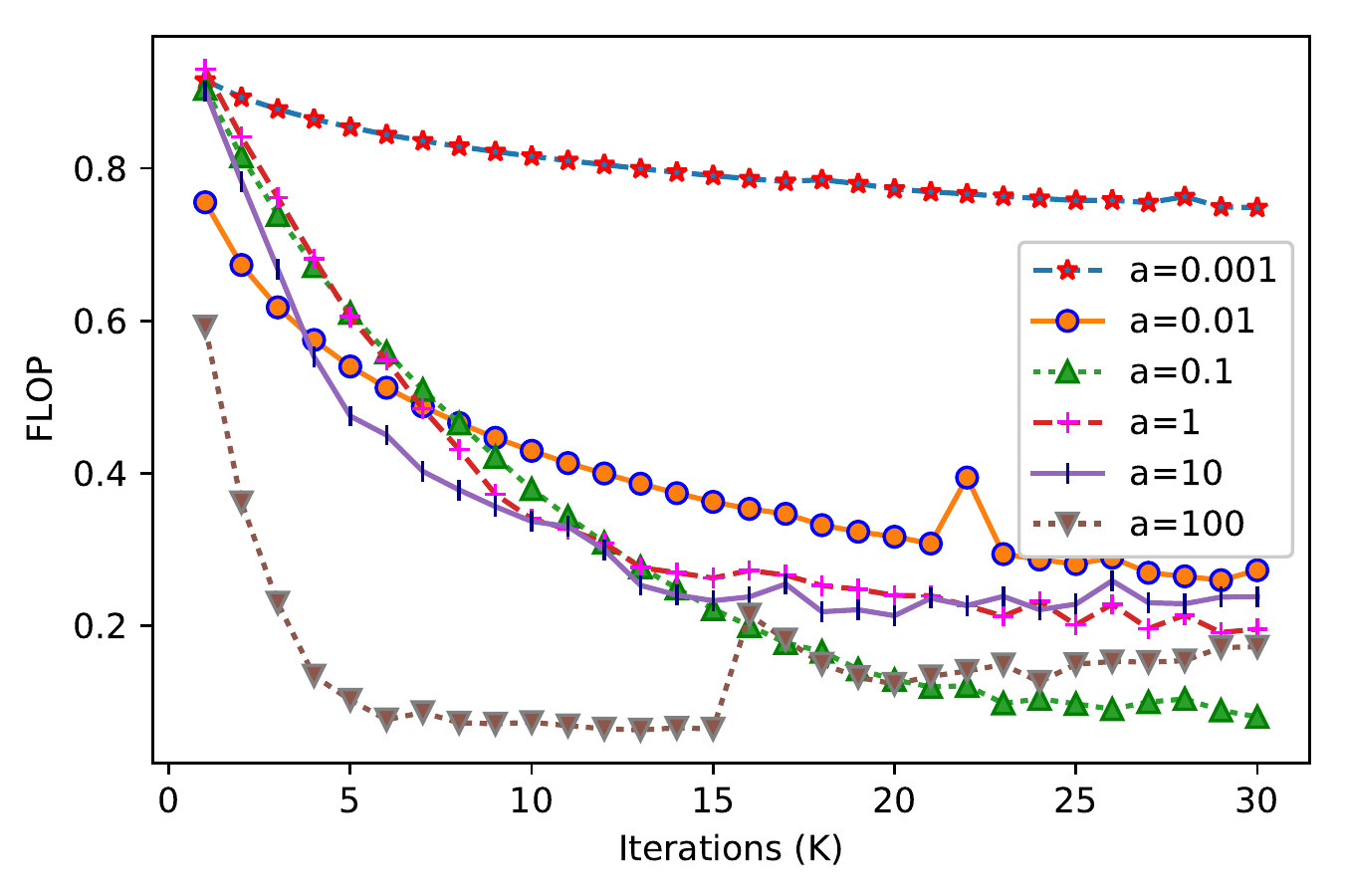}}
	\subfigure[Percentage of parameters]{
		\label{para}
		\includegraphics[width=0.48\textwidth]{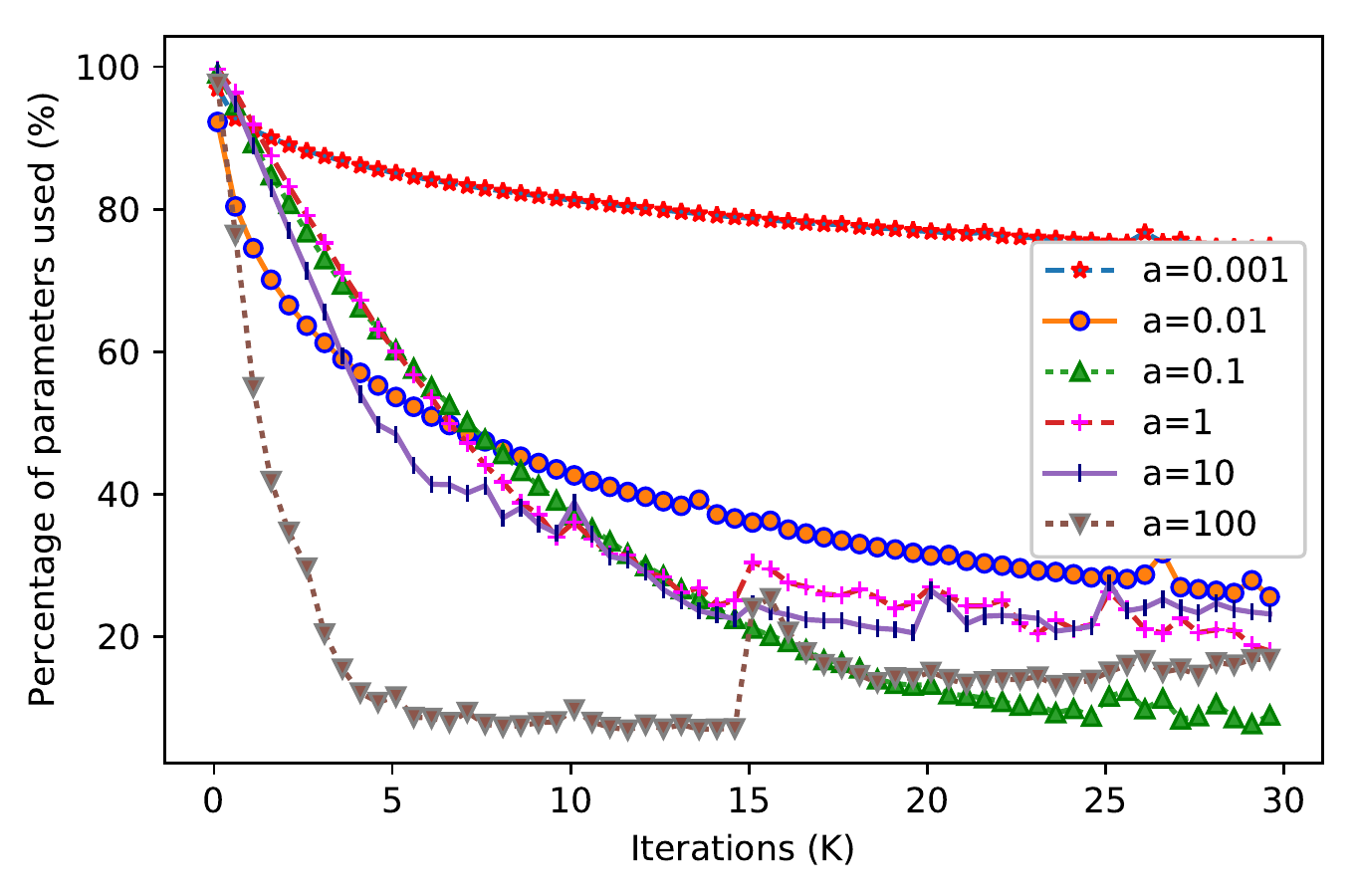}}
	\caption{\textbf{The effect of $a$ on the network.} The prediction accuracy, FLOP, parameters used of the networks with $a$ varying among $\{10^{-3}, 10^{-2}, 10^{-1}, 1, 10, 10^{2}\}$.}
	\label{roleofa}
\end{figure*}

\subsection{Effect of $a$ in integrated T$\ell_1$}
In the experiments, the parameter $a$ in T$\ell_1$ is set in advance. As mentioned previously, when $a$ tends to zero, the T$\ell_1$ approaches the $\ell_0$ norm, while T$\ell_1$ approaches $\ell_1$ when $a$ is close to infinity. In this subsection, we explore the effect of $a$ in integrated T$\ell_1$ by varying the value of $a$ in the range of $\{10^{-3},10^{-2}, 10^{-1}, 1, 10, 10^2\}$ on dataset DIGITS. In all cases, we use a network with one convolutional layer followed by two fully connected layers. For each $a$, we tune other parameters in the model to obtain the best performance.

We display the curves of prediction accuracy, the FLOP and the percentage of parameters used for networks with different values of $a$ in Fig. \ref{accuracy}, \ref{flop} and \ref{para}, respectively. As seen from Fig. \ref{accuracy}, the network with $a=10^{-3}$ achieves the highest accuracy, followed by $a=10^{-2}$ and $a=1$. Furthermore, among networks corresponding to the six values of $a$, the network with $a=10^{-2}$ converges fastest. The network with $a=10^{-1}$ achieves the least number of parameters, while it is slightly worse in accuracy than other compared networks. The FLOP and percentage of parameters of other four settings have little difference, varying from $0.2$ to $0.3$.

\begin{table*}[!t]
	\centering\small	
	\caption{Sparsity-promoting performance of each regularizer}
	\label{inter}{
		\begin{tabular}{ccc}
			\toprule
			Regularizer& Neurons removed& Sparsity of connections($\%$)\\
			\midrule
			Group sparsity & $64$ & $51.60\%$\\
			\midrule
			T$\ell_1$ & $0$ & $61.25\%$\\
			\midrule
			Integrated T$\ell_1$ & $12$ & $76.88\%$\\
			\bottomrule
		\end{tabular}
	}
\end{table*}

\subsection{Interpretation of the regularizers}
In this subsection, to quantitatively demonstrate the sparsity-inducing capacity of group sparsity, T$\ell_1$ and integrated T$\ell_1$, we study the final layers of networks with these three regularizers on dataset DIGITS. The final layer of the complete network is equipped with $128$ neurons. Sparsity-promoting performance of these regularizers are listed in Table \ref{inter}. As we can see from the table, group sparsity is able to remove $64$ neurons and $51.60\%$ connections, indicating that group sparsity can only achieve neuron-level sparsity and the remaining connections are still dense. The T$\ell_1$ cannot remove neurons, but it can remove connections efficiently. Although the network regularized by integrated T$\ell_1$ only removes $12$ neurons, it can achieve $76.88\%$ sparsity, which means that the integrated T$\ell_1$ is able to promote both neuron-level and connection-level sparsity simultaneously.

\section{Conclusion}
In this work, we introduced a new sparsity-inducing regularization called integrated transformed $\ell_1$ regularizer, where a group sparsity regularizer explores structural information of neural networks and removes redundant neurons and a transformed $\ell_1$ norm enforces sparsity between network connections. We verify the performance of our regularizer on several public datasets. Experimental results demonstrate the effectiveness of the proposed regularizer, when compare it with three prominent baselines. 

There is still some research that we wish to explore in the future. To begin with, in this paper, we only verify the effect of integrated T$\ell_1$ on convolutional neural networks. In the future, we intend to test integrated T$\ell_1$ on other neural network architectures. In addition, we plan to use other regularizers to replace $\ell_2$ in group sparsity to group variables. By doing this, we wish to propose a single regularizer that can remove both redundant neurons and connections simultaneously.

\section*{Acknowledgments}
This work was supported by National Natural Science Foundation of China under Grant No.$11671379$ .

\bibliography{manuscript}
\bibliographystyle{elsarticle-num-names}
\biboptions{sort&compress}
\end{document}